\documentclass{article}

\usepackage[preprint]{neurips_2026}

\usepackage[utf8]{inputenc}
\usepackage[T1]{fontenc}
\usepackage{microtype}
\usepackage{amsmath}
\usepackage{amssymb}
\usepackage{amsfonts}
\usepackage{booktabs}
\usepackage{xspace}
\usepackage{graphicx}
\usepackage{enumitem}
\usepackage[table]{xcolor}
\usepackage{url}
\usepackage{hyperref}
\usepackage[capitalize,noabbrev]{cleveref}
\usepackage{comment}
\usepackage{tabularx}
\usepackage{pifont}
\usepackage{nicefrac}
\usepackage{wrapfig}
\usepackage{caption}

\newlength{\experimenttablevsep}
\newcommand{\experimenttablewidth}{0.40\textwidth}

\hypersetup{
  colorlinks=true,
  linkcolor={red},
  citecolor={orange!70!black},
  urlcolor={blue!80!black},
}

\newcommand{\slmmux}{SLM-MUX\xspace}

\newcommand{\slm}{SLM\xspace}
\newcommand{\slms}{SLMs\xspace}
\newcommand{\mathbench}{MATH-500\xspace}
\newcommand{\gsm}{GSM8K\xspace}
\newcommand{\gpqa}{GPQA\xspace}
\newcommand{\mmlupro}{MMLU-Pro\xspace}
\newcommand{\acc}{\mathrm{acc}}
\newcommand{\eps}{\varepsilon}

\definecolor{takeawaygray}{gray}{0.20}
\definecolor{mutedgreen}{RGB}{74, 124, 89}

\title{OrchSLM: Probing the Dynamics of Small Language Model Orchestration}
\author{
Chengxi Zhang\thanks{Equal contribution; order decided by a coin flip.}\\
Harvard\\
\texttt{chengxizhang@g.harvard.edu}
\And
Yu Yao\footnotemark[1]\\
MIT\\
\texttt{yu\_yao@mit.edu}
}
\begin{document}
\maketitle
\begin{abstract}

Although large language models (LLMs) have demonstrated remarkable capabilities, their reliance on cloud-scale infrastructure poses fundamental challenges for deployment in agentic pipelines, including latency, privacy, connectivity, and substantial computational cost. Small language models (SLMs) offer a compelling alternative: recent studies suggest that many repetitive and narrowly scoped subtasks in agentic workloads may be better served by specialized SLMs than by monolithic LLMs. However, the limited capacity and context windows of SLMs can constrain long-horizon reasoning and interaction-heavy orchestration strategies such as iterative verification and debate. This motivates a complementary, non-interactive paradigm in which heterogeneous SLMs independently generate candidate solutions and a router orchestrates their cached samples without further model interaction. To further understand the mechanisms of such orchestration, we introduce OrchSLM, a routing framework that unifies existing non-interactive orchestration methods and exposes their underlying design choices as controllable parameters. Using OrchSLM as a systematic probe, we reveal how orchestration behavior emerges from diverse knobs, including the task structure, model-pool composition, and multi-agent consensus.

\end{abstract}

\section{Introduction}
Large language models have demonstrated remarkable progress across
reasoning, coding, tool use, and increasingly complex agentic tasks
\citep{openai2023gpt4,
touvron2023llama2,grattafiori2024llama3,
deepseekai2025deepseekr1, snell2024scalingllmtesttimecompute}.
Scaling model capacity and inference-time computation has further extended
their ability to solve challenging problems
\citep{kaplan2020scaling,hoffmann2022training,
lightman2023lets,brown2024large,snell2024scaling}.
However, these capabilities come with substantial computational and deployment
costs, particularly when language models must be invoked repeatedly as
components of agentic pipelines. As an alternative, small language models are increasingly emerging as low-cost,
deployable, and heterogeneous reasoning engines
\citep{wang2024comprehensivesurveysmalllanguage,belcak2025small}.
Their modest computational footprint makes it practical to query multiple
models at inference time, raising the possibility of treating a collection
of \slms as a pool of complementary specialists. However, most orchestration
strategies developed for larger language models rely heavily on interaction:
models exchange natural-language messages, critique intermediate solutions,
synthesize one another's responses, or verify candidate answers before a
final decision is made
\citep{wang2024mixtureofagentsenhanceslargelanguage,
du2023improvingfactualityreasoninglanguage,
lifshitz2025multiagentverificationscalingtesttime}.
For \slms, such interaction can be costly relative to the models themselves
and relies on deliberation and self-correction capabilities that remain
limited
\citep{Taubenfeld_2024,
huang2024largelanguagemodelsselfcorrect,
liu2023lostmiddlelanguagemodels,fu2025cache}.

This motivates a complementary, \textit{non-interactive}\footnote{
We use \textit{non-interactive} to mean that models do not communicate with
one another or condition their generations on other models' outputs.
Each model generates independently, and cross-model information is combined
only by the router after generation, without consuming additional context
tokens for inter-model communication.
} setting: each model independently generates a set of cached samples, and final answers are obtained by combining the resulting answer distributions at inference time. Despite its simplicity, a systematic understanding of such non-interactive orchestration still remain unknown. Existing methods make
fundamentally different implicit assumptions about what constitutes useful evidence.
Query-level routers and cascades choose a model before generation, typically
based on predicted expertise, cost, or preference
\citep{chen2023frugalgptuselargelanguage,
ong2025routellmlearningroutellms,
chen2024routerdcquerybasedrouterdual}.
Voting methods instead treat agreement across independently generated answers
as evidence
\citep{wang2023selfconsistency,li2024agentsneed}.
Routing-style routing estimates within-model consistency and favors answers
from models that repeatedly support the same prediction
\citep{wang2025slmmux}.
These approaches expose distinct signals (\textit{expertise},
\textit{self-consistency}, and \textit{mutual agreement}) but entangle them
through different routing rules.

To understand these signals is important because each can be informative in
one regime and misleading in another. A strong expert may provide a reliable
default answer, yet expert selection discards corroborating evidence from the
rest of the pool. Majority agreement can reveal shared evidence, yet a
collection of weaker models may reinforce the same plausible error
\citep{li2024agentsneed,wang2025slmmux}. High self-consistency can indicate
confidence, but it can also arise from a model that repeatedly makes the same
mistake
\citep{wang2023selfconsistency,wang2025slmmux}. Even the composition of the
model pool introduces a nontrivial tradeoff: adding another model may increase
the probability that some model produces the correct answer while
simultaneously introducing additional incorrect support that makes the answer
harder to identify. Consequently, orchestration quality is governed not by
the number of available models or the amount of agreement alone, but by how
{expertise}, {agreement}, and {pool composition}
interact with the structure of the task. This raises the central question of our work:
\textit{what governs the dynamics of small language model orchestration?}
Rather than searching for a single universally optimal router, we seek to
isolate the mechanisms that determine when heterogeneous \slms complement one
another, when collective agreement provides useful evidence, and when adding
more models instead makes routing harder.

To probe these tradeoffs systematically, we decompose non-interactive
multi-\slm orchestration along three research axis.
\textit{(1) Whom should we trust?}
How should authority be distributed across models?
\textit{(2) How should evidence combine?}
How should multiple beliefs form a collective decision?
\textit{(3) When does agreement matter?}
When does consensus provide evidence beyond individual confidence?
Existing routing methods implicitly make different choices along these
dimensions. Majority voting weights models equally, accumulates support
across the pool, and relies on mutual agreement. Single-best routing places
all trust in a fixed expert. Router-style favors the model exhibiting the
strongest self-consistency, while cross-support treats probability mass
assigned by other models to a candidate answer as question-specific
evidence. We make these choices explicit as controllable axes, allowing
existing routers and their intermediate variants to be studied within a
common design space.

We instantiate this design space as a generalized routing family in which
common orchestration methods emerge as exact parameter settings, and use it
as a controlled probe of orchestration dynamics. Across \mathbench, \gsm, \gpqa, and \mmlupro with a heterogeneous model
pool, our systematic analysis reveals distinct dynamics of multi-\slm
orchestration: effective routing emerges from the interplay between task
structure, model expertise, pool composition, and agreement patterns,
rather than from any single source of evidence.

\textbf{Contributions.}
(1) We formulate non-interactive multi-\slm orchestration as a controlled
design space organized around three coupled mechanisms: \textit{whom to
trust}, \textit{how to aggregate evidence}, and \textit{what form of
consistency should constitute confidence}. (2) We introduce \textsc{OrchSLM}, a generalized routing family that exposes
these mechanisms as controllable axes and recovers majority voting,
self-consistency routing, cross-support, expert-weighted routing, and the
single-best limit as specific configurations within a common framework. (3) We use \textsc{OrchSLM} to systematically probe the dynamics of
multi-\slm orchestration across tasks and model pools. Our experiments reveal
how expertise, agreement, and pool composition jointly determine routing
behavior, including a persistent gap between \textit{oracle coverage} and
\textit{routable support}: more models can make the correct answer available
without making it easier to identify.

\section{Preliminaries and Methodology}
\label{sec:method}
\vspace{-10pt}

\begin{figure}[h]
    \centering
    \includegraphics[width=1\linewidth]{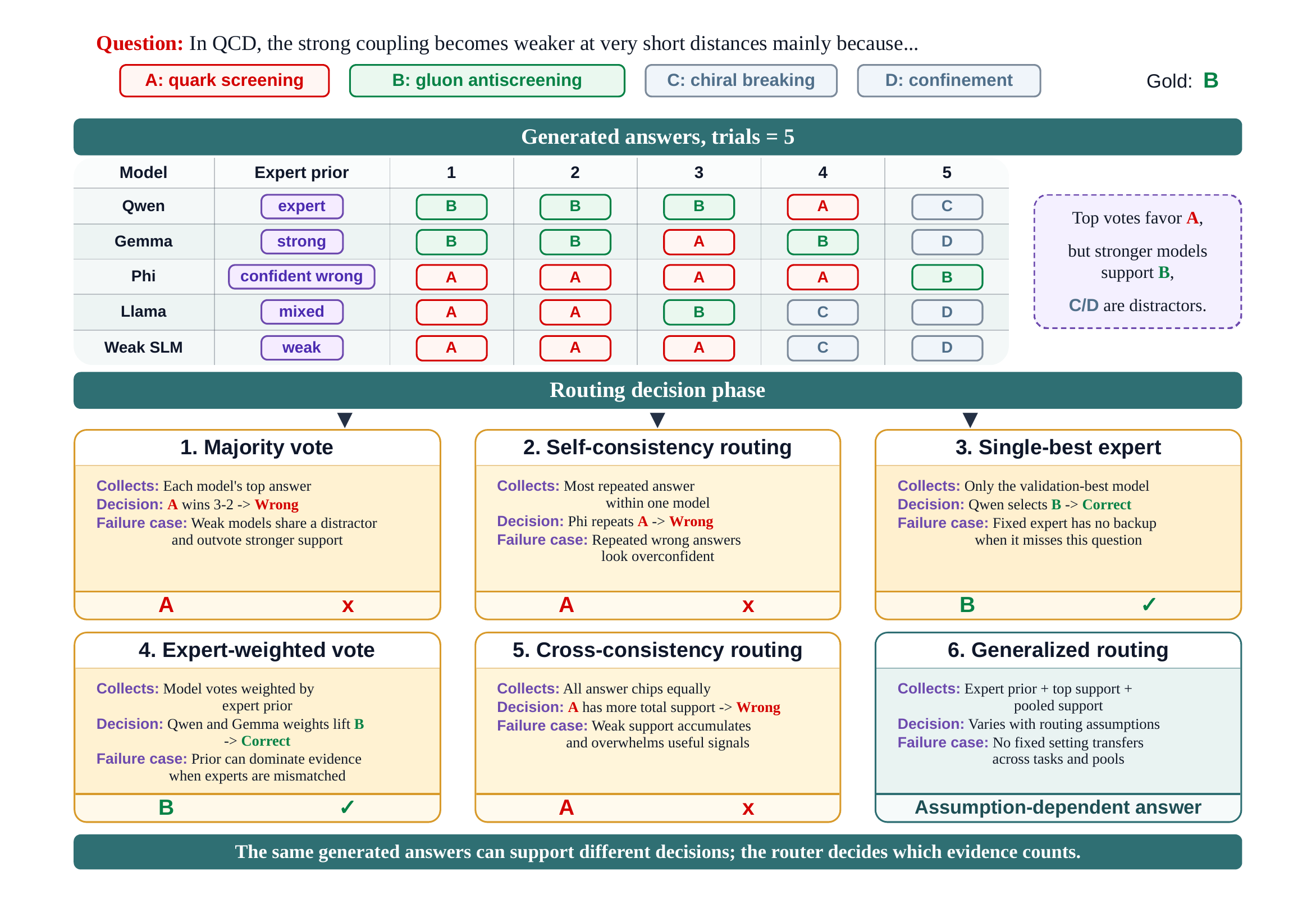}
    \vspace{-30pt}
    \caption{Different routers can pick different answers from the same cached outputs.}
    \label{fig:baseline-failures}
\end{figure}

\textbf{Motivation.}
Orchestration is a problem of collective decision-making \citep{arrow1951social}:
given multiple imperfect and heterogeneous reasoners, how should their
individual beliefs be transformed into a collective judgment?
More voices do not necessarily imply more knowledge, agreement does not
necessarily imply truth, and expertise does not necessarily imply correctness
on every instance. The challenge is therefore not merely to combine model
outputs, but to determine {when individual authority should prevail,
when evidence should accumulate, and when agreement should be trusted}. We study this problem through three fundamental questions.
\textit{Whom should we trust?}
Models differ in expertise, raising a tension between democratic aggregation
and expert authority.
\textit{How should evidence combine?}
Support may derive its strength from accumulation across many independent
sources, or from the conviction of a particularly reliable one.
\textit{When does agreement matter?}
Repeated belief can reflect genuine confidence, but consensus can also
amplify shared errors; its value therefore depends on who agrees and how
their evidence is formed. These questions expose three distinct but coupled mechanisms underlying
multi-\slm orchestration: \textit{trust}, \textit{aggregation}, and
\textit{consistency}. We next formalize them as controllable axes of a
generalized routing family, allowing different orchestration strategies to
be studied within a common framework.

\subsection{Evaluation of Orchestration}
\label{sec:eval}

How should collective evidence become a decision, and whose evidence
actually shapes that decision?
We evaluate orchestration through these two complementary perspectives.
The first concerns \textit{decision}: we formalize how trust, aggregation,
and agreement determine which answer is selected.
The second concerns \textit{influence}: we distinguish between models that
are heard by the router and models whose presence actually improves its
decision.
Together, these views characterize not only what an orchestration system
decides, but also how that collective decision emerges.
Additional derivations and implementation details are provided in
Appendix~\S\ref{app:math}.

\textbf{From samples to beliefs.}
For a question $x$, each model $i$ independently generates $K$ samples.
After answer extraction and normalization, these samples induce an
empirical distribution over candidate answers:
\begin{equation}
P_{ia}(x)
=
\frac{1}{K}
\sum_{k=1}^{K}
\mathbf{1}[y_{ik}=a].
\end{equation}
Here, $P_{ia}(x)$ represents the empirical support that model $i$ assigns
to answer $a$. Let
$a_i^\star=\arg\max_a P_{ia}(x)$
denote model $i$'s most frequent answer, with ties broken deterministically,
and define
$M_{ia}=\mathbf{1}[a=a_i^\star]$.
Thus, $P_{ia}$ captures the model's full answer distribution, while
$M_{ia}$ identifies the answer on which its own belief concentrates most
strongly.

\begin{table*}[h]
\begingroup
\scriptsize
\refstepcounter{table}
\caption{Common routers mapped to $(\alpha,\kappa,\mathrm{Agg})$ in the generalized family.}
\label{tab:method-axes}

\setlength{\tabcolsep}{2pt}
\renewcommand{\arraystretch}{1.2}

\resizebox{\linewidth}{!}{%
\begin{tabular}{p{0.18\linewidth}p{0.39\linewidth}p{0.08\linewidth}p{0.12\linewidth}p{0.24\linewidth}}
\toprule
\textbf{Method} & \textbf{Examples} & \textbf{Trust} & \textbf{Aggregation} & \textbf{Position in family} \\
\midrule
Expert selection
& Query routing, cascades, model routers
& Expert
& Single
& $(1,\infty,-)$ \\

Majority voting
& Self-consistency voting and multi-agent majority
& Average
& Pooled sum
& $(0,0,-)$ \\

Weighted voting
& Cross-validation weighted probabilistic ensemble
& Expert
& Pooled sum
& $(0,\kappa>0,-)$ \\

Self-consistency
& \slmmux{}-style answer-level routing
& Average
& Max
& $(1,0,\mathrm{Max})$ \\

Weighted consistency
& Validation-aware consistency routing
& Expert
& Max
& $(1,\kappa>0,\mathrm{Max})$ \\

Cross consistency
& Answer-level cross-confidence and support variants
& Average
& Max + Pooled
& $(0<\alpha<1,0,\mathrm{Max})$ \\

Generalized routing
& This generalized family
& Tunable
& Tunable
& $([0,1],[0,\infty],\{\mathrm{Sum},\mathrm{Max}\})$ \\

\bottomrule
\end{tabular}%
}
\endgroup
\end{table*}

\begin{wrapfigure}{r}{0.55\linewidth}
    \centering
    \vspace{-5pt}
    \includegraphics[width=\linewidth]{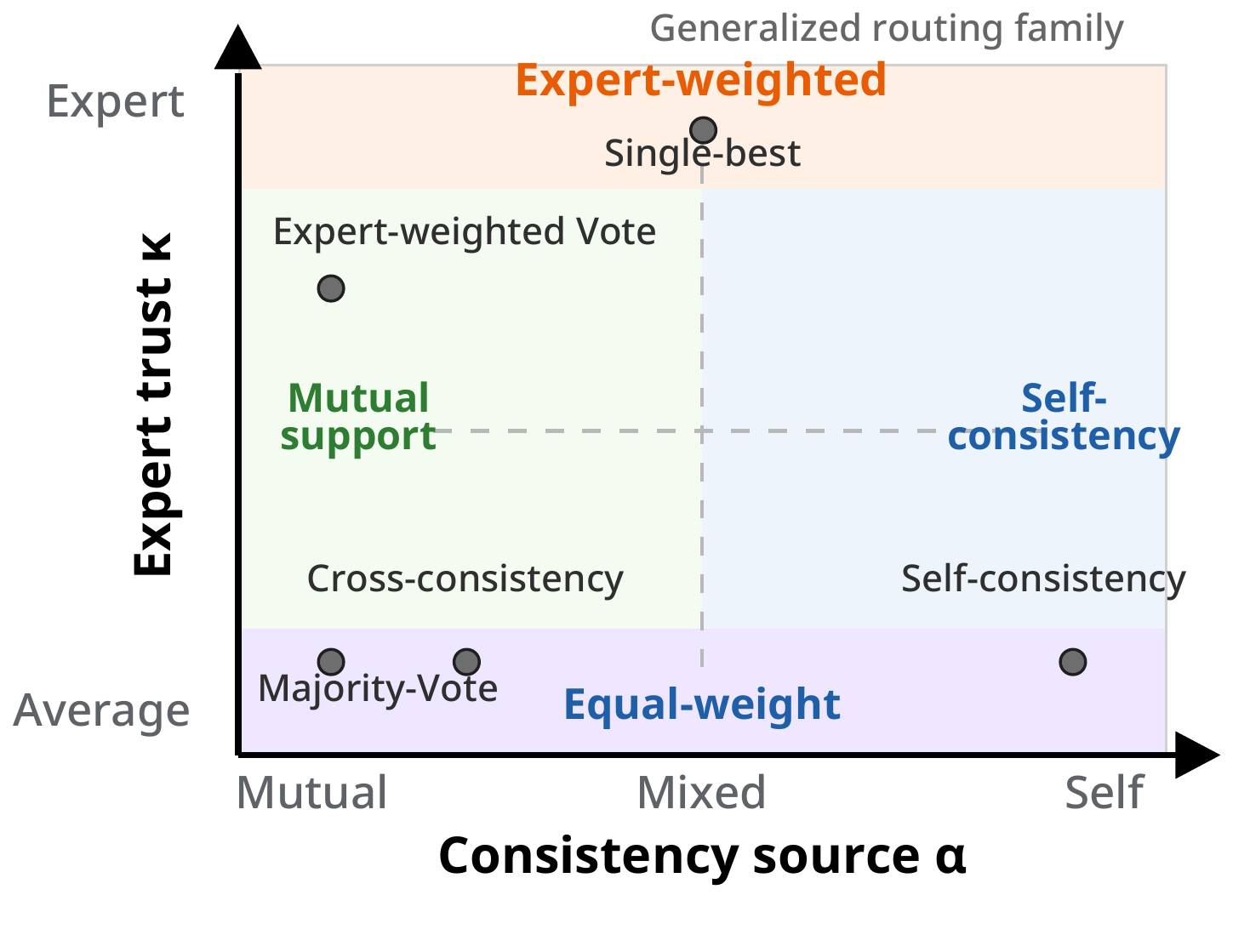}
    \vspace{-8pt}
    \caption{Existing routers as fixed corners of the $(\alpha,\kappa,\mathrm{Agg})$
    generalized family; axes are defined in Section~\ref{sec:eval}.}
    \label{fig:regime-map}
    \vspace{-10pt}
\end{wrapfigure}
\textbf{Trust: whose evidence should count?}
Models need not be equally reliable.
Let $\acc_i$ denote an expert prior for model $i$, estimated from a
validation split or supplied as a fixed prior score.
We translate these priors into normalized trust weights:
\begin{equation}
r_i(\kappa)
=
\frac{\max(\acc_i,\epsilon)^\kappa}
{\sum_j \max(\acc_j,\epsilon)^\kappa}.
\end{equation}
The parameter $\kappa$ determines how concentrated authority is across
the model pool.
At $\kappa=0$, every model receives equal weight, corresponding to a
democratic prior over expertise.
As $\kappa$ increases, authority shifts toward models with stronger prior
performance; in the limit $\kappa\rightarrow\infty$, it concentrates on
the validation-best model.
Thus, $\kappa$ continuously interpolates between collective equality and
expert authority.

\textbf{Aggregation and agreement: how should evidence combine?}
For each candidate answer $a$, we define the routing score
\begin{equation}
W(a)
=
\alpha\,
\mathrm{Agg}_i
\left[
M_{ia}r_i(\kappa)P_{ia}(x)
\right]
+
(1-\alpha)
\sum_i r_i(\kappa)P_{ia}(x),
\end{equation}
and select $\hat a=\arg\max_a W(a)$.
Here, $\mathrm{Agg}$ controls how evidence combines across models:
Sum accumulates support from multiple models, whereas Max retains the
strongest supporter. The parameter $\alpha$ controls the form of
agreement, interpolating between support for each model's preferred
answer and pooled support over its full answer distribution.
Together, $(\kappa,\mathrm{Agg},\alpha)$ specify whom to trust, how to
combine their evidence, and which form of agreement to use.

\textbf{Influence: being heard versus being helpful.}
The routing score determines which answer is selected, but does not tell
us whether each model helps make that selection correct. We therefore
measure model influence through attribution and contribution.

For a dataset $D$, let $\hat a_t$ be the routed answer for question $t$.
We define the attribution of model $i$ as
\begin{equation}
\mathrm{Atr}(i)
=
\frac{1}{|D|}
\sum_{t\in D}
P^{(t)}_{i,\hat a_t}.
\end{equation}
Atr measures how strongly a model supports the answers selected by the
router. A model with high Atr is frequently \textit{heard}, but its
support need not improve accuracy.

To measure whether a model is helpful, we remove it and rerun the router.
We define its counterfactual contribution as
\begin{equation}
\mathrm{Ctr}(i)
=
\mathrm{Acc}(\text{full pool})
-
\mathrm{Acc}(\text{pool without }i).
\end{equation}
Equivalently, if $\hat a_t^{(-i)}$ is the routed answer without model
$i$, then
\begin{equation}
\mathrm{Ctr}(i)
=
\frac{1}{|D|}
\sum_{t\in D}
\left(
\mathbf{1}[\hat a_t=y_t]
-
\mathbf{1}[\hat a_t^{(-i)}=y_t]
\right).
\end{equation}


\section{Experiments}
\label{sec:experiments}

\noindent\textbf{Setup.}
All routers operate on the same cached generations from a seven-model pool
(Qwen2.5-7B-Instruct, Meta-Llama-3.1-8B-Instruct, Llama-3.1-Tulu-3-8B-DPO,
DeepSeek-Math-7B-Instruct, Gemma-2-9B-IT, Phi-3.5-mini-instruct,
Ministral-8B-Instruct-2410) across four benchmarks: \mathbench
(competition math, one dominant expert), \gsm (saturated arithmetic),
\gpqa (hard, balanced science), and \mmlupro (broad multi-topic science). Implementation details are at Appendix~\S\ref{app:math}.

\subsection{Scaling up Task Structures}

\begin{figure*}[htbp]
    \centering
    \includegraphics[width=1.0\textwidth]{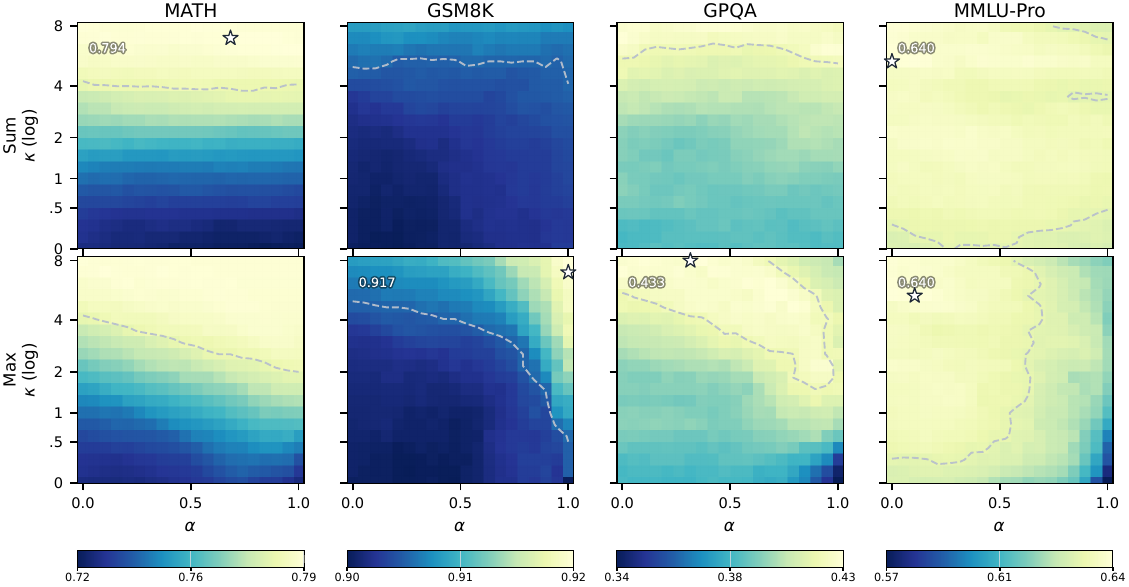}
    \caption{Accuracy heatmaps over $(\alpha, \kappa)$ for Sum and Max aggregation on the full pool. Dashed contour: cells within $1$ pp of the best score; stars: best per task.}
    \label{fig:phase-heatmaps}
\end{figure*}
\begin{figure*}[!ht]
\centering

\begin{minipage}[c]{0.57\textwidth}
\centering

\includegraphics[width=\linewidth]{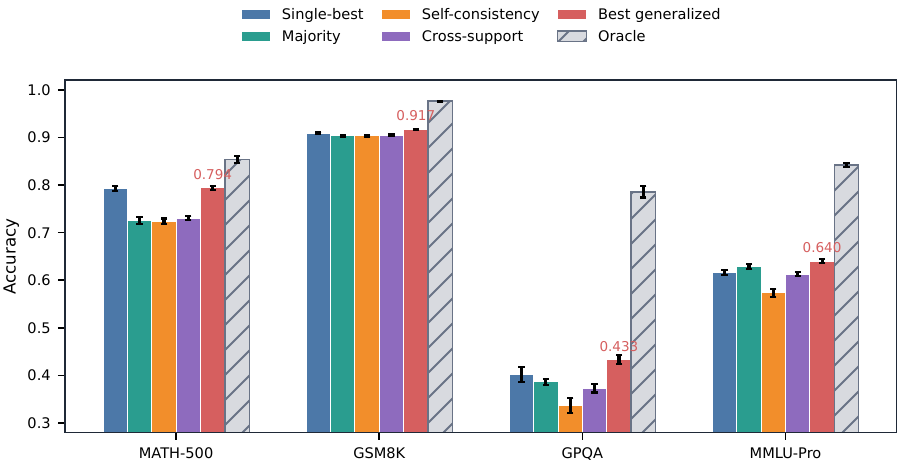}
\captionof{figure}{Router generalizations.}
\label{fig:main-baselines}

\end{minipage}
\hfill
\begin{minipage}[c]{0.38\textwidth}
\centering
\scriptsize

\captionof{table}{Optimal-point behavior under the OrchSLM parameter scan.
The selected optima emphasize expert trust on the two math benchmarks, while
the science benchmarks rely more on mixed or mutual support.}
\label{tab:step45-scorecard}

\setlength{\tabcolsep}{2.5pt}
\renewcommand{\arraystretch}{0.95}

\resizebox{\linewidth}{!}{%
\begin{tabular}{lllll}
\toprule
Task & Trust & Consistency & Aggregation & Regime \\
\midrule
MATH-500 & Expert & Mixed & Tie & Expert \\
GSM8K & Expert & Self & Max & Saturated \\
GPQA & Expert & Mixed & Tie & Support \\
MMLU-Pro & Mixed & Mutual & Tie & Pooled \\
\bottomrule
\end{tabular}%
}

\end{minipage}

\end{figure*}

The best routing strategy varies across tasks.
\Cref{fig:main-baselines} compares four cache-only routing strategies
with the oracle upper bound. On \mathbench, single-best routing is
near-optimal, indicating that routing is largely determined by the
strongest expert. On \gsm, several strategies perform similarly. In
contrast, \gpqa and \mmlupro retain substantial gaps to the oracle,
suggesting that the presence of a correct answer does not make it easy
to identify. These differences motivate us to study routing as a
task-dependent decision rather than fixing a single strategy across
tasks. We vary the three components of the generalized router:
the trust concentration $\kappa$, the aggregation operator
$\mathrm{Agg}\in\{\mathrm{Sum},\mathrm{Max}\}$, and the balance $\alpha$
between top-answer and pooled support.
\Cref{fig:phase-heatmaps} shows the resulting routing landscape, and
\Cref{tab:step45-scorecard} summarizes the best regime for each task.
We next isolate the effect of each component.

\begin{wraptable}{r}{\experimenttablewidth}
\centering
\scriptsize
\caption{Experts vs average models. Rel. $\Delta$ is the relative change from the best equal-weight cell ($\kappa=0$) to the best full-sweep cell.}
\label{tab:step45-expert-trust}

\resizebox{\linewidth}{!}{%
\begin{tabular}{lrrr}
\toprule
Task & Equal & Expert & Rel. $\Delta$ \\
\midrule
MATH-500 & 0.7338 & 0.7938 & +8.2\% \\
GSM8K & 0.9052 & 0.9166 & +1.3\% \\
GPQA & 0.3909 & 0.4328 & +10.7\% \\
MMLU-Pro & 0.6304 & 0.6402 & +1.6\% \\
\bottomrule
\end{tabular}%
}
\end{wraptable}
\textbf{Trust concentration.} Expert weighting has the largest effect on routing performance.
\Cref{tab:step45-expert-trust} compares the best setting with
$\kappa>0$ against the best equal-weight setting with $\kappa=0$.
Concentrating trust consistently improves routing across all four
tasks. On math, it prevents weaker models from diluting the strongest
expert. On the harder science tasks, it reduces the influence of
unreliable support. Thus, deciding which models to trust is the primary routing decision.

\textbf{Aggregation.} The choice between Sum and Max has a smaller effect.
As shown in \Cref{tab:step45-aggregation}, \gsm clearly favors Max,
while the best Sum and Max settings are close on the other three tasks.
This is consistent with \gsm being a saturated task, where selecting
the strongest expert-supported answer is sufficient. In the remaining
tasks, changing how support is aggregated matters less than changing
which models receive weight.

\begin{wraptable}{r}{\experimenttablewidth}
\centering
\scriptsize
\caption{Majority vs coalition support, vs strongest supporter. Rel. $\Delta$ is the relative change from Sum to Max aggregation.}
\label{tab:step45-aggregation}

\resizebox{\linewidth}{!}{%
\begin{tabular}{lrrr}
\toprule
Task & Sum & Max & Rel. $\Delta$ \\
\midrule
MATH-500 & 0.7938 & 0.7936 & 0.0\% \\
GSM8K & 0.9089 & 0.9166 & +0.8\% \\
GPQA & 0.4318 & 0.4328 & +0.2\% \\
MMLU-Pro & 0.6398 & 0.6402 & +0.1\% \\
\bottomrule
\end{tabular}%
}
\end{wraptable}

\textbf{Agreement.}
The usefulness of self-consistency and mutual support depends on which
models the router trusts.
\Cref{tab:step45-consistency} compares the two signals by setting
$\alpha=1$ for top-answer support and $\alpha=0$ for pooled support
under a fixed $\kappa$. With equal model weights, the two signals are
similar and neither consistently dominates. After expert weighting,
their roles separate: math favors self-consistent expert predictions,
whereas the broad science task benefits more from pooled support.
Agreement is therefore informative only relative to the models
providing it.

\subsection{Scaling up Routers}

\begin{figure*}[!th]
\centering

\begin{minipage}[t]{0.5\textwidth}
\vspace{0pt}

\centering
\scriptsize
\captionof{table}{Self-consistency vs mutual agreement. Pooled-only uses
$\alpha=0$; top-only uses $\alpha=1$. Rel. $\Delta$ is the relative change
from pooled-only to top-only routing.}
\label{tab:step45-consistency}

\resizebox{\linewidth}{!}{%
\begin{tabular}{lrrrrrr}
\toprule
 & \multicolumn{3}{c}{All $\kappa$} & \multicolumn{3}{c}{$\kappa=0$} \\
\cmidrule(lr){2-4}\cmidrule(lr){5-7}
Task & Pooled & Top & Rel. $\Delta$ & Pooled & Top & Rel. $\Delta$ \\
\midrule
MATH-500 & 0.7934 & 0.7936 & 0.0\% & 0.7246 & 0.7242 & -0.1\% \\
GSM8K & 0.9086 & 0.9166 & +0.9\% & 0.9032 & 0.9050 & +0.2\% \\
GPQA & 0.4318 & 0.4308 & -0.2\% & 0.3864 & 0.3879 & +0.4\% \\
MMLU-Pro & 0.6398 & 0.6350 & -0.8\% & 0.6282 & 0.6270 & -0.2\% \\
\bottomrule
\end{tabular}%
}

\scriptsize
\captionof{table}{Oracle coverage is not routable support.}
\label{tab:oracle-gap}

\begin{tabular}{lrrr}
\toprule
Benchmark & Oracle & Best generalized & Gap \\
\midrule
MATH-500 & 0.8536 & 0.7938 & 0.0598 \\
GSM8K & 0.9760 & 0.9166 & 0.0594 \\
GPQA & 0.7854 & 0.4328 & 0.3525 \\
MMLU-Pro & 0.8420 & 0.6402 & 0.2018 \\
\bottomrule
\end{tabular}

\end{minipage}
\hfill
\begin{minipage}[t]{.45\textwidth}
\vspace{-25pt}

\centering
\includegraphics[width=\linewidth]{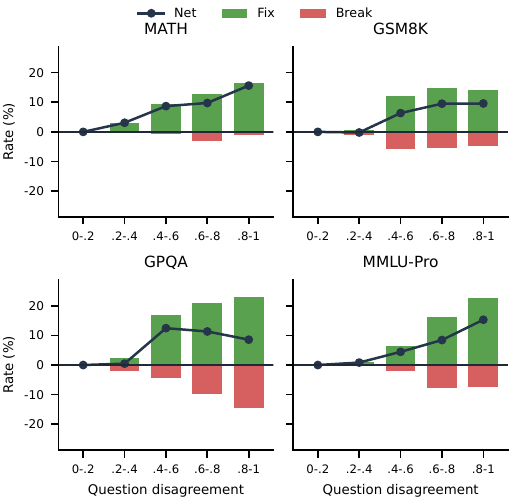}

\captionof{figure}{Per-bin fixes (green) and breaks (red) when switching
self-consistency to the best generalized cell, with the net rate (line).}
\label{fig:fix-break}

\end{minipage}

\end{figure*}
\textbf{Oracle coverage and routability.}
Higher oracle coverage does not necessarily translate into higher routed
accuracy.
\Cref{tab:oracle-gap} compares the oracle upper bound with the best
generalized routing result. The gap is small on the math tasks but
substantially larger on \gpqa and \mmlupro. On these science benchmarks,
a correct answer is often present in the model pool but cannot be
reliably identified from the cached predictions. Increasing coverage
alone is therefore insufficient; the router must also distinguish
useful evidence from competing support.

\textbf{Disagreement.}
Disagreement creates both opportunities and risks for routing.
\Cref{fig:fix-break} groups questions by the level of disagreement among
the seven models and measures how generalized routing changes the
predictions of self-consistency. As disagreement increases, the router
corrects more self-consistency errors, but it also changes more correct
predictions into incorrect ones. The largest net benefit occurs at
moderate disagreement, where alternative evidence is available without
being dominated by conflicting support.

\begin{figure*}[h]
    \centering
    \begin{minipage}[t]{0.49\linewidth}
        \centering
        \includegraphics[width=\linewidth]{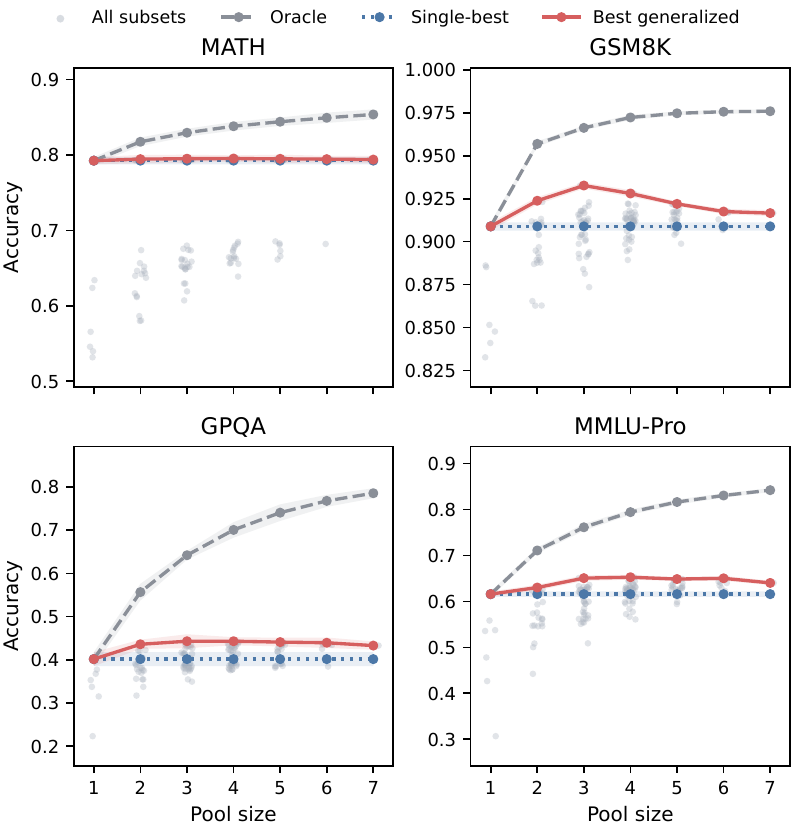}
    \end{minipage}
    \hfill
    \begin{minipage}[t]{0.49\linewidth}
        \centering
        \includegraphics[width=\linewidth]{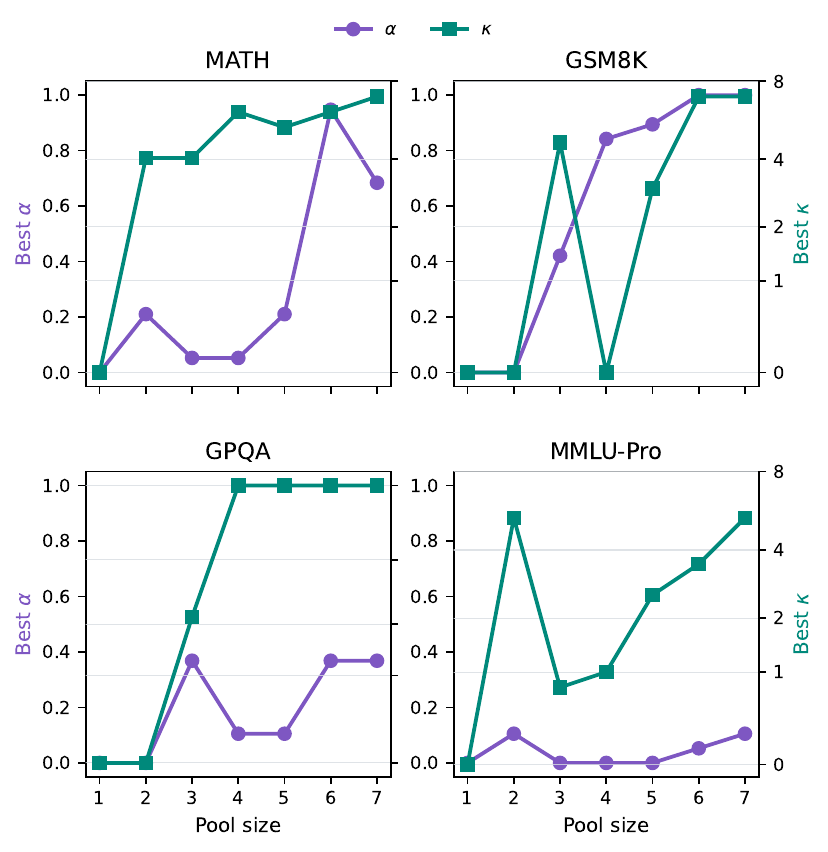}
    \end{minipage}
    \caption{Pool-size scan. Left: routed accuracy vs.\ size (gray: each subset; lines: oracle, single-best, best generalized). Right: best-cell $(\alpha,\kappa)$ shifts with pool size.}
    \label{fig:pool-size}
\end{figure*}

\textbf{Pool composition.}
Adding more models does not necessarily improve routed accuracy.
As shown in \Cref{fig:pool-size} (left), oracle coverage continues to
increase with pool size, while routed accuracy typically peaks with
three or four models. Additional models can introduce correct answers
that increase oracle coverage, but they can also introduce stable
incorrect support that makes those answers harder to identify.
Consequently, the quality of the pool depends not only on the answers
it contains, but also on the support patterns it creates. Pool composition also changes the preferred routing strategy.
\Cref{fig:pool-size} (right) shows that the optimal $(\alpha,\kappa)$
shifts as models are added or removed. Smaller curated pools generally
require less trust concentration because the remaining models are
already more reliable. Model selection and answer routing are therefore
coupled decisions rather than independent stages.

\subsection{Being Heard Is Not Being Helpful}

\begin{figure*}[h]
    \centering
    \includegraphics[width=0.98\linewidth]{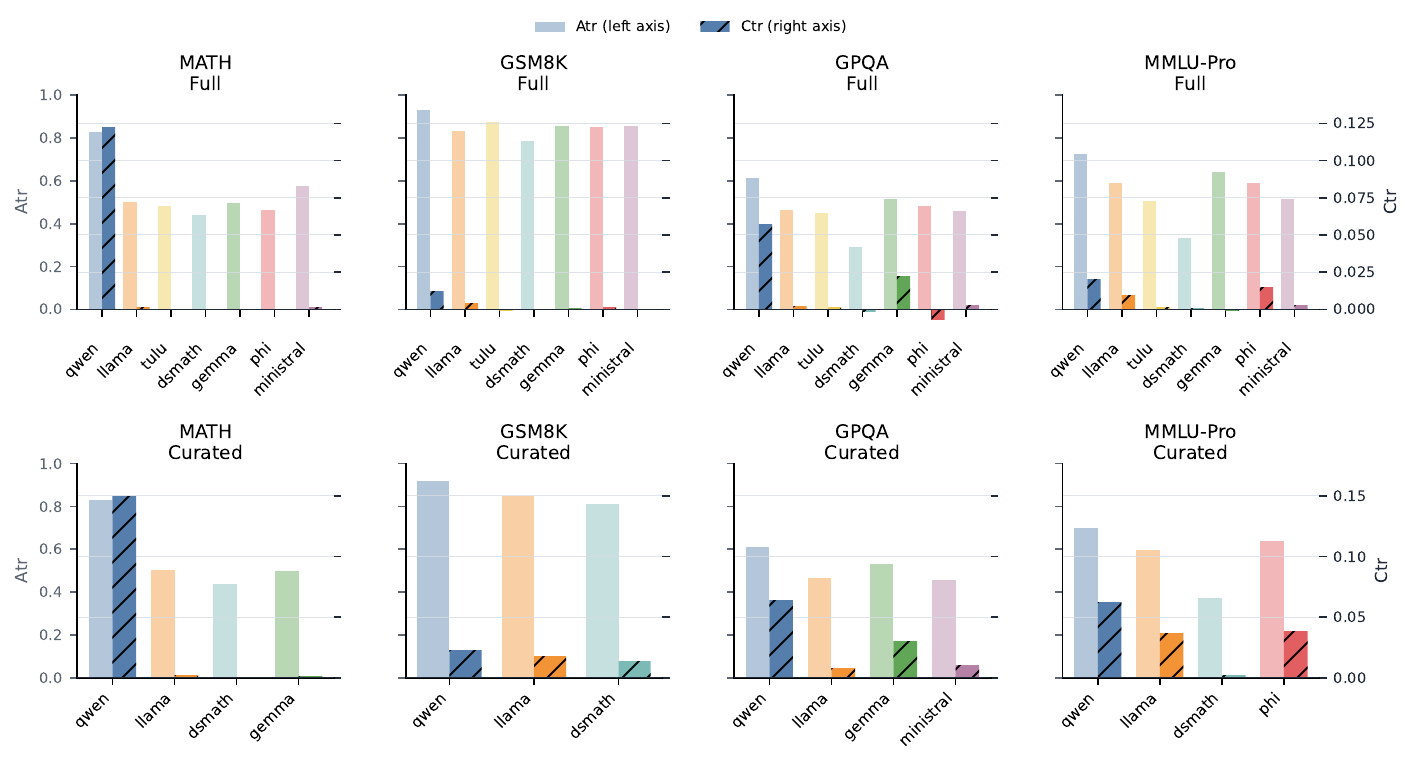}
    \caption{Per-model Atr (heard) vs.\ Ctr (helpful) under each canonical router.}
    \label{fig:atr-ctr}
\end{figure*}
Attribution and contribution capture different aspects of model
influence. Atr measures how much probability mass a model assigns to
the router's selected answer, whereas Ctr measures how routed accuracy
changes when that model is removed. A model can therefore receive high
attribution without making the router more accurate. \Cref{fig:atr-ctr} shows that the two measures behave differently across
tasks. On math, the strongest expert has both high Atr and positive Ctr:
the model that most strongly supports the routed answers is also the
one that improves them. On \gpqa and \mmlupro, however, several models
receive substantial Atr while having near-zero or negative Ctr. Their
predictions frequently agree with the router's decisions, but removing
them does not hurt accuracy and can sometimes improve it. Pool curation reduces this mismatch. After removing models with
unhelpful support, attribution becomes more concentrated on models with
positive counterfactual contribution. This explains why expert weighting
and pool selection can improve routing on the harder science tasks:
effective orchestration depends not only on how much support a model
provides, but also on whether that support is useful for selecting the
correct answer. Per-model results are reported in
\Cref{tab:atr-ctr-by-model}, with an accepted-support decomposition in
\Cref{tab:app-step4-e328-attribution}.

\section{Related Work}
\label{sec:related-work}

\textbf{Small Language Models.}
Small language models (SLMs) trade capacity for lower latency, memory, and
deployment cost \citep{abdin2024phi3,liu2024mobilellm}.  Recent gains at this
scale come from curated and synthetic data, knowledge distillation,
scale-specific architectures, and targeted post-training
\citep{wang2024comprehensivesurveysmalllanguage,abdin2024phi3,
allal2025smollm2,gemmateam2024gemma2,liu2024mobilellm}.  This efficiency
supports repeated or specialized calls on devices and in agentic systems
\citep{abdin2024phi3,liu2024mobilellm,belcak2025small,wang2025slmmux}, while
chain-of-thought, repeated sampling, and test-time scaling improve reasoning
at inference time
\citep{wei2022cot,wang2023selfconsistency,
snell2024scalingllmtesttimecompute,muennighoff2025s1simpletesttimescaling}.
However, weak self-correction and long-context use constrain interaction-heavy
orchestration \citep{huang2024largelanguagemodelsselfcorrect,
liu2023lostmiddlelanguagemodels}.  SLM-MUX instead uses independent sampling,
confidence-based selection, and model-subset search \citep{wang2025slmmux};
OrchSLM studies routing within this non-interactive setting.

\textbf{Multi-Agent and Compound AI Systems.}
Multi-agent and compound AI systems organize multiple model calls into a
single solution or decision.  At the system level, routing and allocation
determine which calls are made, while prompt and workflow search configure how
they are used
\citep{chen2023frugalgptuselargelanguage,
ong2025routellmlearningroutellms,chen2024routerdcquerybasedrouterdual,
chen2025llmselector,poon2025onlinemultillm,
khattab2023dspycompilingdeclarativelanguage,
opsahlong2024optimizinginstructionsdemonstrationsmultistage,
saadfalcon2025archonarchitecturesearchframework,
zhang2025aflowautomatingagenticworkflow}.  Given a call configuration, systems
differ in whether outputs interact during generation.  Interactive systems
condition later calls on earlier outputs through debate, critique, synthesis,
or verification
\citep{du2023improvingfactualityreasoninglanguage,
wang2024mixtureofagentsenhanceslargelanguage,
lifshitz2025multiagentverificationscalingtesttime}.  MoA variants refine this
pattern through proposer quality and diversity, sparse communication, and
residual, attention, or memory-based flow
\citep{li2025selfmoa,li2025smoa,xie2025rmoa,wen2026attentionmoa,
ping2026remmoa}.  Non-interactive systems instead sample independently and
aggregate only after generation \citep{wang2025slmmux}.  OrchSLM focuses on
this post-generation decision.

\textbf{Ensembling and Test-Time Inference.}
Test-time inference improves fixed models or model pools through additional
sampling, search, verification, and aggregation
\citep{snell2024scalingllmtesttimecompute,
zhang2025surveytesttimescalinglarge}.  In its non-interactive branch, the
remaining problem is how to select or combine independently generated
candidates.  Existing rules range from sample voting and verifier-based
selection to confidence-weighted consensus, multi-generator and reward-model
ensembles, and reasoning-structure aggregation
\citep{wang2023selfconsistency,li2024agentsneed,lightman2023lets,
yao2026roundtablepolicy,song2025ctts,parulekar2026reasoningconsensus,
wang2025slmmux}.  These rules can be organized along three axes: model authority
(equal or quality-weighted), evidence aggregation (accumulated support or the
strongest supporter), and confidence source (self-consistency or mutual
agreement).  Attribution methods add a separate question: whether a model
merely supports a decision or changes it when removed
\citep{ribeiro2016why,lundberg2017unified,koh2017understanding}.
OrchSLM parameterizes these choices and separates answer availability from
routability and received support from counterfactual contribution.

\section{Conclusion}
We study SLM orchestration through three decisions: whom to trust, how
to aggregate evidence, and which form of agreement to use. The best
choice depends on both the task and the model pool. Expert weighting is
the strongest routing factor, larger pools do not necessarily improve
accuracy, and agreement is useful only when it comes from helpful
models. These results suggest that effective orchestration requires
jointly selecting the models and deciding how their evidence should be
combined.

\section*{Limitations}
Our study is limited to cache-only, non-interactive orchestration and
benchmarks with extractable answers. Extending the framework to
interactive or open-ended settings would require additional generations
and semantic equivalence mechanisms. Expert priors also require
validation data or an estimate of model quality. Finally, while our
all-subset analysis explores the pool design space, we use split-selected
evaluation to test pool selection on held-out data.

A natural next step is to use routing signals to decide when additional
interaction, verification, or debate is worth the compute. The same
signals could support adaptive pool construction, selecting models and
routing strategies at the question or domain level. Extending OrchSLM
to open-ended generation would further require semantic matching and
partial-credit evaluation while preserving interpretable influence
diagnostics.

\section*{Disclosure of Generative AI Usage}
We used large language models to assist with writing clarity, grammar, and experimental code development. All final wording, technical claims, code, results, figures’ scientific content, and conclusions were reviewed, validated, and produced by the authors.

\bibliographystyle{plainnat}
\bibliography{reference}

\clearpage
\clearpage
\appendix
\appendix

\part*{Appendices}

\textbf{Overview.} The first sections provide protocol details,
full-pool analyses, pool-composition results, robustness checks, and
Atr/Ctr diagnostics for the seven-model experiments. The final sections
collect additional mechanism and sensitivity tables that support the
same interpretation from smaller or alternative analysis settings.

\section{Mathematical Interpretation}
\label{app:math}

Generalized routing score contains two sources of evidence:

\noindent \textbf{Top-answer support.}
The first term,
\begin{equation}
\mathrm{Agg}_i
\left[
M_{ia}\,r_i(\kappa)\,P_{ia}(x)
\right],
\end{equation}
only counts model $i$'s support for answer $a$ if $a$ is that model's own
top answer. This term measures how strongly the models that actually
select $a$ support it. The operator $\mathrm{Agg}$ determines how this
support is combined across models:
\begin{itemize}
    \item $\mathrm{Sum}$ adds support from all models whose top answer is
    $a$:
    \begin{equation}
    \mathrm{Sum}_i[z_i]=\sum_i z_i.
    \end{equation}
    This corresponds to coalition-style evidence, where several models
    can jointly support the same answer.

    \item $\mathrm{Max}$ keeps only the strongest individual supporter:
    \begin{equation}
    \mathrm{Max}_i[z_i]=\max_i z_i.
    \end{equation}
    This corresponds to strongest-supporter evidence, where one highly
    confident or highly trusted model can dominate.
\end{itemize}

\noindent \textbf{Pooled answer support.}
The second term,
\begin{equation}
\sum_i r_i(\kappa)P_{ia}(x),
\end{equation}
counts every model's probability mass on answer $a$, even if $a$ is not
that model's top answer. This term measures cross-model or mutual support
for $a$.

The interpolation parameter $\alpha\in[0,1]$ controls which type of
evidence dominates. When $\alpha=1$, routing relies only on top-answer
support. When $\alpha=0$, routing relies only on pooled support. Thus,
$\kappa$ controls whom to trust, $\mathrm{Agg}$ controls whether evidence
comes from a coalition or the strongest supporter, and $\alpha$ controls
whether confidence comes from top-answer self-consistency or pooled
mutual agreement.

\section{Experimental Details}

\subsection{Stability and Robustness Check}

The conclusions in experiments section persist under held-out pool selection
(\cref{app:pool-composition}, \cref{tab:app-split-selected}),
subsample-replicate variability (\cref{tab:robustness},
\cref{tab:app-step4-e313-replicate-stability}), and full-cache $K{=}50$
evaluation (\cref{tab:app-k50-phase,tab:app-keff}). Prior-transform,
tie-breaking, exact cross-support, and fix/break checks are reported in
\cref{app:robustness-ablations}.

\subsection{Protocol Details and Summary Extraction}

The main experiments use K20R10: for each question and model, the
evaluation samples 20 answers from a K=50 cache and repeats this
deterministic subsampling protocol 10 times. Tables report means over
replicates unless otherwise noted. The main generalized sweep uses fixed
expert priors to analyze routing assumptions. The all-subset pool scans
study how pool composition changes routed accuracy; split-selected
evaluation tests the same selection idea under held-out evaluation. The
small main-text summary tables extract views of the same full-pool sweep
to answer the three routing questions directly.

\begin{table*}[h!]
\centering
\small
\caption{Task-specific routing regimes from the full-pool generalized sweep. The table summarizes broad preferred regions; near-best plateau statistics are reported separately.}
\label{tab:task-regimes}
\begin{tabular}{lrrrll}
\toprule
Benchmark & Best acc. & $\alpha$ & $\kappa$ & Agg. & Interpretation \\
\midrule
MATH-500 & 0.7938 & 0.6842 & 7.0171 & Sum & single-best-like expert routing \\
GSM8K & 0.9166 & 1.0000 & 7.0171 & Max & expert self-consistency \\
GPQA & 0.4328 & 0.3684 & 8.0000 & Max & expert + support-aware routing \\
MMLU-Pro & 0.6402 & 0.1053 & 5.3617 & Max & expert-weighted pooled support \\
\bottomrule
\end{tabular}
\end{table*}

\clearpage
\subsection{Full-Pool Routing Decisions}

This section contains the full-pool analyses behind the main
three-decision analysis. The tables report plateau width, single-best
gaps, no-expert-prior ablations, pooled-term and top-term ablations,
Sum-vs-Max comparisons, and disagreement-bin behavior. These numbers are
the table counterpart of the main heatmap and regime discussion.

\begin{table*}[h!]
\centering
\scriptsize
\caption{Full-pool plateau summary. Broad near-best regions support interpreting best cells as regimes with stable neighboring cells.}
\label{tab:app-plateau-full}
\begin{tabular}{llrrrrrrr}
\toprule
Benchmark & Best acc. & Best params & Cells @0.25pp & Area @0.25pp & Cells @0.5pp & Area @0.5pp & Cells @1pp & Area @1pp \\
\midrule
GPQA & 0.4328 & Max a=0.3684 k=8.0 & 19 & 0.0238 & 78 & 0.0975 & 158 & 0.1975 \\
GSM8K & 0.9166 & Max a=1.0 k=7.0171 & 15 & 0.0187 & 39 & 0.0488 & 210 & 0.2625 \\
MATH-500 & 0.7938 & Sum a=0.6842 k=7.0171 & 137 & 0.1713 & 199 & 0.2487 & 273 & 0.3412 \\
MMLU-Pro & 0.6402 & Max a=0.1053 k=5.3617 & 38 & 0.0475 & 213 & 0.2662 & 566 & 0.7075 \\
\bottomrule
\end{tabular}
\end{table*}

\begin{table*}[h!]
\centering
\scriptsize
\caption{Expert-prior axis: gap between the best generalized full-pool cell and the single-best limit.}
\label{tab:app-single-best-gap}
\begin{tabular}{lllll}
\toprule
Benchmark & Best gen. & Single-best & Gap & Best $\kappa$ \\
\midrule
MATH-500 & 0.7938 & 0.7924 & 0.0014 & 7.0171 \\
GSM8K & 0.9166 & 0.9089 & 0.0077 & 7.0171 \\
GPQA & 0.4328 & 0.4015 & 0.0313 & 8.0000 \\
MMLU-Pro & 0.6402 & 0.6158 & 0.0244 & 5.3617 \\
\bottomrule
\end{tabular}
\end{table*}

\begin{table*}[h!]
\centering
\scriptsize
\caption{No-expert-prior comparison on the full seven-model pool, read from the $\kappa=0$ best cell in the full-pool phase diagram.}
\label{tab:app-no-expert-prior}
\begin{tabular}{lrrrrrr}
\toprule
Benchmark & Best $\kappa{=}0$ & Std & Agg. & $\alpha$ & Full-family best & Gap \\
\midrule
MATH-500 & 0.7338 & 0.0055 & Max & 0.7368 & 0.7938 & -0.0600 \\
GSM8K & 0.9052 & 0.0013 & Max & 0.8947 & 0.9166 & -0.0114 \\
GPQA & 0.3909 & 0.0120 & Sum & 0.6316 & 0.4328 & -0.0419 \\
MMLU-Pro & 0.6304 & 0.0069 & Sum & 0.2632 & 0.6402 & -0.0098 \\
\bottomrule
\end{tabular}
\end{table*}

\begin{table*}[h!]
\centering
\scriptsize
\caption{Top-answer-term removal ablation on the full seven-model pool. Setting $\alpha=0$ keeps only pooled answer support.}
\label{tab:app-alpha0}
\begin{tabular}{llllll}
\toprule
Benchmark & Best $\alpha{=}0$ & Agg. & $\kappa$ & Full-family best & Gap \\
\midrule
GPQA & 0.4318 & Sum & 7.0171 & 0.4328 & -0.0010 \\
GSM8K & 0.9086 & Sum & 8.0000 & 0.9166 & -0.0080 \\
MATH-500 & 0.7934 & Sum & 8.0000 & 0.7938 & -0.0004 \\
MMLU-Pro & 0.6398 & Sum & 5.3617 & 0.6402 & -0.0004 \\
\bottomrule
\end{tabular}
\end{table*}

\begin{table*}[h!]
\centering
\scriptsize
\caption{Pooled-term removal ablation on the full seven-model pool. Setting $\alpha=1$ removes pooled mutual support.}
\label{tab:app-alpha1}
\begin{tabular}{llllll}
\toprule
Benchmark & Best $\alpha{=}1$ & Agg. & $\kappa$ & Full-family best & Gap \\
\midrule
GPQA & 0.4308 & Sum & 6.1416 & 0.4328 & -0.0020 \\
GSM8K & 0.9166 & Max & 7.0171 & 0.9166 & 0.0000 \\
MATH-500 & 0.7936 & Sum & 6.1416 & 0.7938 & -0.0002 \\
MMLU-Pro & 0.6350 & Sum & 5.3617 & 0.6402 & -0.0052 \\
\bottomrule
\end{tabular}
\end{table*}

\begin{table*}[h!]
\centering
\scriptsize
\caption{Sum/Max summary. The aggregation switch measures whether support is distributed across models or concentrated in the strongest supporter.}
\label{tab:app-sum-max-summary}
\begin{tabular}{lllll}
\toprule
benchmark & benchmark\_label & pool\_size & mean\_sum\_minus\_max & frac\_sum\_wins \\
\midrule
gpqa & GPQA & 1.0000 & 0.0000 & 0.0000 \\
gpqa & GPQA & 2.0000 & 0.0000 & 0.0000 \\
gpqa & GPQA & 3.0000 & -0.0002 & 0.2857 \\
gpqa & GPQA & 4.0000 & 0.0004 & 0.5143 \\
gpqa & GPQA & 5.0000 & -0.0002 & 0.4286 \\
gpqa & GPQA & 6.0000 & -0.0006 & 0.2857 \\
gpqa & GPQA & 7.0000 & -0.0010 & 0.0000 \\
gsm8k & GSM8K & 1.0000 & 0.0000 & 0.0000 \\
gsm8k & GSM8K & 2.0000 & 0.0000 & 0.0000 \\
gsm8k & GSM8K & 3.0000 & -0.0031 & 0.2857 \\
gsm8k & GSM8K & 4.0000 & -0.0024 & 0.2000 \\
gsm8k & GSM8K & 5.0000 & -0.0053 & 0.0952 \\
gsm8k & GSM8K & 6.0000 & -0.0049 & 0.0000 \\
gsm8k & GSM8K & 7.0000 & -0.0077 & 0.0000 \\
math500 & MATH-500 & 1.0000 & 0.0000 & 0.0000 \\
math500 & MATH-500 & 2.0000 & 0.0000 & 0.0000 \\
math500 & MATH-500 & 3.0000 & 0.0000 & 0.1429 \\
math500 & MATH-500 & 4.0000 & 0.0000 & 0.2571 \\
math500 & MATH-500 & 5.0000 & 0.0001 & 0.3810 \\
math500 & MATH-500 & 6.0000 & 0.0002 & 0.5714 \\
\bottomrule
\end{tabular}
\end{table*}

\begin{table*}[h!]
\centering
\scriptsize
\caption{Full-pool best generalized accuracy by pair-disagreement bin.}
\label{tab:app-step4-e307-disagreement-bins}
\begin{tabular}{llrr}
\toprule
Benchmark & Disagreement bin & $N$ & Accuracy \\
\midrule
GPQA & (-0.01, 0.2] & 80 & 0.5250 \\
GPQA & (0.2, 0.4] & 214 & 0.4346 \\
GPQA & (0.4, 0.6] & 587 & 0.4889 \\
GPQA & (0.6, 0.8] & 795 & 0.4050 \\
GPQA & (0.8, 1.01] & 304 & 0.3717 \\
GSM8K & (-0.01, 0.2] & 8882 & 0.9895 \\
GSM8K & (0.2, 0.4] & 1894 & 0.9018 \\
GSM8K & (0.4, 0.6] & 1513 & 0.7072 \\
GSM8K & (0.6, 0.8] & 515 & 0.6738 \\
GSM8K & (0.8, 1.01] & 386 & 0.4560 \\
MATH-500 & (-0.01, 0.2] & 1700 & 0.9859 \\
MATH-500 & (0.2, 0.4] & 715 & 0.9259 \\
MATH-500 & (0.4, 0.6] & 610 & 0.8361 \\
MATH-500 & (0.6, 0.8] & 605 & 0.7008 \\
MATH-500 & (0.8, 1.01] & 1370 & 0.5088 \\
MMLU-Pro & (-0.01, 0.2] & 659 & 0.8998 \\
MMLU-Pro & (0.2, 0.4] & 845 & 0.7988 \\
MMLU-Pro & (0.4, 0.6] & 1262 & 0.7013 \\
MMLU-Pro & (0.6, 0.8] & 1056 & 0.5284 \\
MMLU-Pro & (0.8, 1.01] & 1178 & 0.4160 \\
\bottomrule
\end{tabular}
\end{table*}

\clearpage
\subsection{Pool Composition and Structure}
\label{app:pool-composition}

This section tests whether the model pool itself changes the routing
problem. The tables compare pools selected for absolute accuracy with
pools selected for gain over the self-consistency router, classify
scanned pools by routing regime, relate structural pool metrics to
routing gains, and evaluate split-selected pool choice under held-out
evaluation.

\begin{table*}[h!]
\centering
\scriptsize
\caption{Pool-composition targets. For each benchmark we report the pool with highest absolute accuracy and the pool with largest gain over the self-consistency router.}
\label{tab:app-step4-e302-pool-targets}
\setlength{\tabcolsep}{3pt}
\renewcommand{\arraystretch}{0.95}

\resizebox{\textwidth}{!}{%
\begin{tabular}{lllrrrrrll}
\toprule
Benchmark & Target & Pool & Size & Best gen. & Std & $\Delta$ vs SLM & $\Delta$ vs SB & Params & Regime \\
\midrule
GPQA & best acc. & qwen+llama+gemma+ministral & 4 & 0.4429 & 0.0093 & 0.0212 & 0.0414 & Sum a=0.1053 k=8.0 & high-kappa expert/single-best-like \\
GPQA & best gain & qwen+tulu+dsmath+phi & 4 & 0.4333 & 0.0182 & 0.1101 & 0.0318 & Sum a=1.0 k=3.4968 & mixed support-aware \\
GSM8K & best acc. & qwen+llama+dsmath & 3 & 0.9327 & 0.0018 & 0.0060 & 0.0238 & Sum a=0.4211 k=4.6669 & high-kappa expert/single-best-like \\
GSM8K & best gain & qwen+gemma+ministral & 3 & 0.9111 & 0.0021 & 0.0227 & 0.0022 & Max a=0.9474 k=8.0 & no stable gain \\
MATH-500 & best acc. & qwen+llama+dsmath+gemma & 4 & 0.7952 & 0.0053 & 0.0558 & 0.0028 & Max a=0.0526 k=6.1416 & no stable gain \\
MATH-500 & best gain & qwen+llama+tulu+dsmath+gemma+phi+ministral & 7 & 0.7938 & 0.0049 & 0.0696 & 0.0014 & Sum a=0.6842 k=7.0171 & no stable gain \\
MMLU-Pro & best acc. & qwen+llama+dsmath+phi & 4 & 0.6526 & 0.0052 & 0.0652 & 0.0368 & Sum a=0.0 k=1.0014 & low/moderate-alpha Sum coalition/Majority-like \\
MMLU-Pro & best gain & llama+dsmath & 2 & 0.5646 & 0.0078 & 0.0936 & 0.0060 & Sum a=0.0 k=1.5222 & low/moderate-alpha Sum coalition/Majority-like \\
\bottomrule
\end{tabular}%
}
\end{table*}
\begin{table*}[h!]
\centering
\small
\caption{Best absolute pool from the all-subset pool-composition scan. The companion figure shows how accuracy changes with pool size, while this table records the selected model names.}
\label{tab:pool-composition}
\begin{tabular}{p{0.14\linewidth}p{0.42\linewidth}rrr}
\toprule
Benchmark & Best absolute pool & Size & Best acc. & Full-pool best \\
\midrule
MATH-500 & qwen+llama+dsmath+gemma & 4 & 0.7952 & 0.7938 \\
GSM8K & qwen+llama+dsmath & 3 & 0.9327 & 0.9166 \\
GPQA & qwen+llama+gemma+ministral & 4 & 0.4429 & 0.4328 \\
MMLU-Pro & qwen+llama+dsmath+phi & 4 & 0.6526 & 0.6402 \\
\bottomrule
\end{tabular}
\end{table*}

\begin{table*}[h!]
\centering
\scriptsize
\caption{Regime classification over all scanned pools. Fractions are within benchmark.}
\label{tab:app-step4-e303-regime-counts}
\begin{tabular}{llrr}
\toprule
Benchmark & Regime & Pools & Frac. \\
\midrule
GPQA & high-alpha Max self-consistency-like & 8 & 0.063 \\
GPQA & high-kappa expert/single-best-like & 72 & 0.567 \\
GPQA & low/moderate-alpha Sum coalition/Majority-like & 12 & 0.094 \\
GPQA & mixed support-aware & 14 & 0.110 \\
GPQA & no stable gain & 21 & 0.165 \\
GSM8K & high-alpha Max self-consistency-like & 3 & 0.024 \\
GSM8K & high-kappa expert/single-best-like & 74 & 0.583 \\
GSM8K & low/moderate-alpha Sum coalition/Majority-like & 12 & 0.094 \\
GSM8K & mixed support-aware & 4 & 0.031 \\
GSM8K & no stable gain & 34 & 0.268 \\
MATH-500 & high-kappa expert/single-best-like & 47 & 0.370 \\
MATH-500 & low/moderate-alpha Sum coalition/Majority-like & 10 & 0.079 \\
MATH-500 & no stable gain & 70 & 0.551 \\
MMLU-Pro & high-kappa expert/single-best-like & 64 & 0.504 \\
MMLU-Pro & low/moderate-alpha Sum coalition/Majority-like & 44 & 0.346 \\
MMLU-Pro & mixed support-aware & 8 & 0.063 \\
MMLU-Pro & no stable gain & 11 & 0.087 \\
\bottomrule
\end{tabular}
\end{table*}

\begin{table*}[h!]
\centering
\scriptsize
\caption{Structure-predictor correlations between pool metrics and generalized-routing gains.}
\label{tab:app-structure-corr}
\begin{tabular}{llll}
\toprule
benchmark\_label & predictor & response & pearson\_r \\
\midrule
GPQA & pool\_size & delta\_best\_vs\_slm & 0.6762 \\
GPQA & pool\_size & delta\_best\_vs\_single\_best & 0.5491 \\
GPQA & pool\_size & delta\_best\_vs\_majority & 0.3595 \\
GPQA & pool\_size & best\_gen\_acc\_mean & 0.5517 \\
GPQA & oracle\_gap\_mean & delta\_best\_vs\_slm & 0.6899 \\
GPQA & oracle\_gap\_mean & delta\_best\_vs\_single\_best & 0.5402 \\
GPQA & oracle\_gap\_mean & delta\_best\_vs\_majority & 0.3930 \\
GPQA & oracle\_gap\_mean & best\_gen\_acc\_mean & 0.5632 \\
GPQA & wrong\_answer\_agreement\_mean & delta\_best\_vs\_slm & 0.6572 \\
GPQA & wrong\_answer\_agreement\_mean & delta\_best\_vs\_single\_best & 0.5492 \\
GPQA & wrong\_answer\_agreement\_mean & delta\_best\_vs\_majority & 0.3542 \\
GPQA & wrong\_answer\_agreement\_mean & best\_gen\_acc\_mean & 0.5714 \\
GPQA & avg\_pair\_disagreement\_mean & delta\_best\_vs\_slm & 0.5734 \\
GPQA & avg\_pair\_disagreement\_mean & delta\_best\_vs\_single\_best & 0.2224 \\
GPQA & avg\_pair\_disagreement\_mean & delta\_best\_vs\_majority & 0.5663 \\
GPQA & avg\_pair\_disagreement\_mean & best\_gen\_acc\_mean & 0.3251 \\
GPQA & stable\_wrong\_rate\_mean & delta\_best\_vs\_slm & 0.8332 \\
GPQA & stable\_wrong\_rate\_mean & delta\_best\_vs\_single\_best & 0.4150 \\
GPQA & stable\_wrong\_rate\_mean & delta\_best\_vs\_majority & 0.5489 \\
GPQA & stable\_wrong\_rate\_mean & best\_gen\_acc\_mean & 0.4800 \\
GPQA & single\_model\_acc\_mean & delta\_best\_vs\_slm & -0.3758 \\
GPQA & single\_model\_acc\_mean & delta\_best\_vs\_single\_best & 0.4190 \\
GPQA & single\_model\_acc\_mean & delta\_best\_vs\_majority & -0.4097 \\
GPQA & single\_model\_acc\_mean & best\_gen\_acc\_mean & 0.6045 \\
GPQA & single\_model\_acc\_std & delta\_best\_vs\_slm & 0.7475 \\
GPQA & single\_model\_acc\_std & delta\_best\_vs\_single\_best & 0.0559 \\
GPQA & single\_model\_acc\_std & delta\_best\_vs\_majority & 0.7916 \\
GPQA & single\_model\_acc\_std & best\_gen\_acc\_mean & 0.2273 \\
GPQA & single\_model\_acc\_spread & delta\_best\_vs\_slm & 0.8318 \\
GPQA & single\_model\_acc\_spread & delta\_best\_vs\_single\_best & 0.1763 \\
GPQA & single\_model\_acc\_spread & delta\_best\_vs\_majority & 0.7839 \\
GPQA & single\_model\_acc\_spread & best\_gen\_acc\_mean & 0.3234 \\
GSM8K & pool\_size & delta\_best\_vs\_slm & 0.4439 \\
GSM8K & pool\_size & delta\_best\_vs\_single\_best & 0.0014 \\
GSM8K & pool\_size & delta\_best\_vs\_majority & 0.3702 \\
GSM8K & pool\_size & best\_gen\_acc\_mean & 0.5492 \\
GSM8K & oracle\_gap\_mean & delta\_best\_vs\_slm & 0.1643 \\
GSM8K & oracle\_gap\_mean & delta\_best\_vs\_single\_best & 0.7245 \\
GSM8K & oracle\_gap\_mean & delta\_best\_vs\_majority & -0.0527 \\
GSM8K & oracle\_gap\_mean & best\_gen\_acc\_mean & 0.2886 \\
\bottomrule
\end{tabular}
\end{table*}

\begin{table*}[h!]
\centering
\scriptsize
\caption{Split-selected evaluation. Variants differ in whether they select only scorer parameters on the full pool, choose from predefined curated pools, or choose from all subsets.}
\label{tab:app-split-selected}
\begin{tabular}{lp{0.28\linewidth}rrrrrr}
\toprule
Benchmark & Variant & Eval acc. & Eval std & Pool size & Delta vs SC & Delta vs single & Delta vs majority \\
\midrule
GPQA & A\_full\_pool\_select\_scorer & 0.4120 & 0.0309 & 7.0000 & 0.0760 & 0.0054 & 0.0253 \\
GPQA & B\_predefined\_curated\_select\_pool\_scorer & 0.4167 & 0.0281 & 3.6200 & 0.0807 & 0.0101 & 0.0300 \\
GPQA & C\_all\_subset\_select\_pool\_scorer & 0.4075 & 0.0319 & 3.9800 & 0.0715 & 0.0009 & 0.0208 \\
GSM8K & A\_full\_pool\_select\_scorer & 0.9149 & 0.0071 & 7.0000 & 0.0114 & 0.0047 & 0.0112 \\
GSM8K & B\_predefined\_curated\_select\_pool\_scorer & 0.9311 & 0.0066 & 3.0000 & 0.0277 & 0.0210 & 0.0275 \\
GSM8K & C\_all\_subset\_select\_pool\_scorer & 0.9306 & 0.0070 & 3.0600 & 0.0271 & 0.0204 & 0.0269 \\
MATH-500 & A\_full\_pool\_select\_scorer & 0.7868 & 0.0161 & 7.0000 & 0.0677 & -0.0003 & 0.0635 \\
MATH-500 & B\_predefined\_curated\_select\_pool\_scorer & 0.7859 & 0.0154 & 3.4200 & 0.0668 & -0.0011 & 0.0627 \\
MATH-500 & C\_all\_subset\_select\_pool\_scorer & 0.7837 & 0.0157 & 3.3400 & 0.0646 & -0.0034 & 0.0604 \\
MMLU-Pro & A\_full\_pool\_select\_scorer & 0.6300 & 0.0225 & 7.0000 & 0.0577 & 0.0147 & 0.0010 \\
MMLU-Pro & B\_predefined\_curated\_select\_pool\_scorer & 0.6375 & 0.0231 & 4.1200 & 0.0653 & 0.0222 & 0.0086 \\
MMLU-Pro & C\_all\_subset\_select\_pool\_scorer & 0.6339 & 0.0228 & 4.2400 & 0.0617 & 0.0186 & 0.0050 \\
\bottomrule
\end{tabular}
\end{table*}

\clearpage
\subsection{Fix/Break, Robustness, and Ablations}
\label{app:robustness-ablations}

This section stress-tests the main conclusions. The fix/break table
separates questions corrected by generalized routing from questions newly
broken by it, including comparisons against several canonical reference
routers. Corrected significance tests check whether selected-pool gains
survive multiple-baseline correction. Replicate-stability tables
summarize K20R10 variability. The remaining tables vary expert priors,
tie-breaking and invalid-answer handling, exact cross-support bridge
strength, K=50 evaluation, effective sample budget, and method-correlation
structure.

\begin{table*}[h!]
\centering
\scriptsize
\caption{Fix/break summary for full and best-absolute pools. Counts are over all K20R10 question-replicate rows.}
\label{tab:app-step4-e306-fix-break}
\begin{tabular}{lllrrrrrr}
\toprule
Benchmark & Pool & Comparison & $N$ & Fix & Break & Target acc. & Base acc. & Net \\
\midrule
GPQA & full & vs slm & 1980 & 341 & 151 & 0.4328 & 0.3369 & 0.0960 \\
GPQA & full & vs single & 1980 & 155 & 93 & 0.4328 & 0.4015 & 0.0313 \\
GPQA & full & vs Majority & 1980 & 199 & 107 & 0.4328 & 0.3864 & 0.0465 \\
GPQA & best\_abs & vs slm & 1980 & 139 & 97 & 0.4429 & 0.4217 & 0.0212 \\
GPQA & best\_abs & vs single & 1980 & 163 & 81 & 0.4429 & 0.4015 & 0.0414 \\
GPQA & best\_abs & vs Majority & 1980 & 108 & 59 & 0.4429 & 0.4182 & 0.0247 \\
GSM8K & full & vs slm & 13190 & 328 & 150 & 0.9166 & 0.9031 & 0.0135 \\
GSM8K & full & vs single & 13190 & 230 & 128 & 0.9166 & 0.9089 & 0.0077 \\
GSM8K & full & vs Majority & 13190 & 298 & 121 & 0.9166 & 0.9032 & 0.0134 \\
GSM8K & best\_abs & vs slm & 13190 & 150 & 71 & 0.9327 & 0.9267 & 0.0060 \\
GSM8K & best\_abs & vs single & 13190 & 447 & 133 & 0.9327 & 0.9089 & 0.0238 \\
GSM8K & best\_abs & vs Majority & 13190 & 58 & 56 & 0.9327 & 0.9325 & 0.0002 \\
MATH-500 & full & vs slm & 5000 & 384 & 36 & 0.7938 & 0.7242 & 0.0696 \\
MATH-500 & full & vs single & 5000 & 9 & 2 & 0.7938 & 0.7924 & 0.0014 \\
MATH-500 & full & vs Majority & 5000 & 425 & 79 & 0.7938 & 0.7246 & 0.0692 \\
MMLU-Pro & full & vs slm & 5000 & 532 & 200 & 0.6402 & 0.5738 & 0.0664 \\
MMLU-Pro & full & vs single & 5000 & 280 & 158 & 0.6402 & 0.6158 & 0.0244 \\
MMLU-Pro & full & vs Majority & 5000 & 217 & 157 & 0.6402 & 0.6282 & 0.0120 \\
\bottomrule
\end{tabular}
\end{table*}

\begin{table*}[h!]
\centering
\scriptsize
\caption{Selected-pool paired tests with max-stat correction across baselines within each benchmark/pool role.}
\label{tab:app-step4-e311-significance}
\begin{tabular}{lllrr}
\toprule
Benchmark & Pool role & Baseline & $\Delta$ & Corrected $p$ \\
\midrule
GPQA & best\_abs & SLM-MUX\_best\_abs & 0.0212 & 0.009 \\
GPQA & best\_abs & single-best\_best\_abs & 0.0414 & 0.000 \\
GPQA & best\_abs & Majority\_best\_abs & 0.0247 & 0.001 \\
GPQA & best\_abs & cross slice\_best\_abs & 0.0268 & 0.000 \\
GPQA & best\_gap & SLM-MUX\_best\_gap & 0.1101 & 0.000 \\
GPQA & best\_gap & single-best\_best\_gap & 0.0318 & 0.003 \\
GPQA & best\_gap & Majority\_best\_gap & 0.0753 & 0.000 \\
GPQA & best\_gap & cross slice\_best\_gap & 0.0889 & 0.000 \\
GSM8K & best\_abs & SLM-MUX\_best\_abs & 0.0060 & 0.001 \\
GSM8K & best\_abs & single-best\_best\_abs & 0.0238 & 0.000 \\
GSM8K & best\_abs & Majority\_best\_abs & 0.0002 & 0.826 \\
GSM8K & best\_abs & cross slice\_best\_abs & 0.0003 & 0.765 \\
\bottomrule
\end{tabular}
\end{table*}

\begin{table*}[h!]
\centering
\scriptsize
\caption{Replicate stability. Means and standard deviations are over the 10 K20R10 subsampling replicates.}
\label{tab:app-step4-e313-replicate-stability}
\begin{tabular}{lllrr}
\toprule
Benchmark & Pool role & Method & Mean acc. & Std \\
\midrule
GPQA & best\_abs & Majority & 0.4182 & 0.0144 \\
GPQA & best\_abs & Single-Best & 0.4015 & 0.0170 \\
GPQA & best\_abs & SLM-MUX & 0.4217 & 0.0122 \\
GPQA & best\_gap & Majority & 0.3581 & 0.0155 \\
GPQA & best\_gap & Single-Best & 0.4015 & 0.0170 \\
GPQA & best\_gap & SLM-MUX & 0.3232 & 0.0203 \\
GPQA & full & Majority & 0.3864 & 0.0072 \\
GPQA & full & Single-Best & 0.4015 & 0.0170 \\
GPQA & full & SLM-MUX & 0.3369 & 0.0163 \\
GSM8K & best\_abs & Majority & 0.9325 & 0.0009 \\
GSM8K & best\_abs & Single-Best & 0.9089 & 0.0025 \\
GSM8K & best\_abs & SLM-MUX & 0.9267 & 0.0019 \\
GSM8K & full & Majority & 0.9032 & 0.0019 \\
GSM8K & full & Single-Best & 0.9089 & 0.0025 \\
GSM8K & full & SLM-MUX & 0.9031 & 0.0019 \\
MATH-500 & full & Majority & 0.7246 & 0.0077 \\
MATH-500 & full & Single-Best & 0.7924 & 0.0051 \\
MATH-500 & full & SLM-MUX & 0.7242 & 0.0056 \\
MMLU-Pro & full & Majority & 0.6282 & 0.0058 \\
MMLU-Pro & full & Single-Best & 0.6158 & 0.0061 \\
MMLU-Pro & full & SLM-MUX & 0.5738 & 0.0089 \\
\bottomrule
\end{tabular}
\end{table*}

\begin{table*}[h!]
\centering
\small
\caption{Robustness checks. Split-selected results use held-out evaluation with predefined curated pools; K=50 uses all cached samples once. The direction of the regimes persists under both checks.}
\label{tab:robustness}
\begin{tabular}{lrrrr}
\toprule
Benchmark & Split-selected acc. & $\Delta$ vs self-consistency & K=50 best gen. & K=50 self-consistency \\
\midrule
MATH-500 & 0.7859 & +0.0668 & 0.8000 & 0.7240 \\
GSM8K & 0.9311 & +0.0277 & 0.9189 & 0.9052 \\
GPQA & 0.4167 & +0.0807 & 0.4394 & 0.3030 \\
MMLU-Pro & 0.6375 & +0.0653 & 0.6480 & 0.5580 \\
\bottomrule
\end{tabular}
\end{table*}

\begin{table*}[h!]
\centering
\scriptsize
\caption{Expert-prior transform variants. These checks test whether the qualitative regimes depend on a single prior transformation.}
\label{tab:app-prior-variants}

\resizebox{\linewidth}{!}{%
\begin{tabular}{lllllllll}
\toprule
benchmark & benchmark\_label & pool\_role & pool\_id & prior\_variant & best\_acc & best\_agg & best\_kappa & best\_alpha \\
\midrule
math500 & MATH-500 & full & qwen+llama+tulu+dsmath+gemma+phi+ministral & softmax\_log\_accuracy & 0.7938 & Sum & 7.0171 & 0.6842 \\
math500 & MATH-500 & full & qwen+llama+tulu+dsmath+gemma+phi+ministral & power\_accuracy & 0.7938 & Sum & 7.0171 & 0.6842 \\
math500 & MATH-500 & full & qwen+llama+tulu+dsmath+gemma+phi+ministral & centered\_logit & 0.7938 & Sum & 2.1786 & 0.7368 \\
math500 & MATH-500 & full & qwen+llama+tulu+dsmath+gemma+phi+ministral & rank\_based & 0.7936 & Sum & 5.3617 & 0.9474 \\
math500 & MATH-500 & full & qwen+llama+tulu+dsmath+gemma+phi+ministral & uniform & 0.7338 & Max & 0.0000 & 0.7368 \\
gsm8k & GSM8K & full & qwen+llama+tulu+dsmath+gemma+phi+ministral & softmax\_log\_accuracy & 0.9166 & Max & 7.0171 & 1.0000 \\
gsm8k & GSM8K & full & qwen+llama+tulu+dsmath+gemma+phi+ministral & power\_accuracy & 0.9166 & Max & 7.0171 & 1.0000 \\
\bottomrule
\end{tabular}%
}
\end{table*}
\begin{table*}[h!]
\centering
\scriptsize
\caption{Tie-break sensitivity. Tie-breaking changes small details while preserving the main regime interpretation.}
\label{tab:app-tiebreak}
\begin{tabular}{lllllll}
\toprule
benchmark & benchmark\_label & pool\_role & pool\_id & tie\_break & accuracy\_mean & accuracy\_std \\
\midrule
math500 & MATH-500 & full & qwen+llama+tulu+dsmath+gemma+phi+ministral & pooled\_lex & 0.7938 & 0.0049 \\
math500 & MATH-500 & full & qwen+llama+tulu+dsmath+gemma+phi+ministral & lex & 0.7938 & 0.0049 \\
math500 & MATH-500 & full & qwen+llama+tulu+dsmath+gemma+phi+ministral & fixed\_model\_order & 0.7938 & 0.0049 \\
math500 & MATH-500 & full & qwen+llama+tulu+dsmath+gemma+phi+ministral & random & 0.7938 & 0.0049 \\
gsm8k & GSM8K & full & qwen+llama+tulu+dsmath+gemma+phi+ministral & pooled\_lex & 0.9166 & 0.0017 \\
gsm8k & GSM8K & full & qwen+llama+tulu+dsmath+gemma+phi+ministral & lex & 0.9166 & 0.0017 \\
gsm8k & GSM8K & full & qwen+llama+tulu+dsmath+gemma+phi+ministral & fixed\_model\_order & 0.9166 & 0.0017 \\
\bottomrule
\end{tabular}
\end{table*}

\begin{table*}[h!]
\centering
\scriptsize
\caption{Exact cross-support bridge sweep. This sweep studies how the bridge varies with cross-support strength.}
\label{tab:app-exact-bridge}
\begin{tabular}{lllllll}
\toprule
benchmark & benchmark\_label & pool\_role & pool\_id & lambda & accuracy\_mean & accuracy\_std \\
\midrule
math500 & MATH-500 & full & qwen+llama+tulu+dsmath+gemma+phi+ministral & 0.0 & 0.7296 & 0.0043 \\
math500 & MATH-500 & full & qwen+llama+tulu+dsmath+gemma+phi+ministral & 0.16666666666666666 & 0.7304 & 0.0042 \\
math500 & MATH-500 & full & qwen+llama+tulu+dsmath+gemma+phi+ministral & 0.25 & 0.7334 & 0.0059 \\
math500 & MATH-500 & full & qwen+llama+tulu+dsmath+gemma+phi+ministral & 0.5 & 0.7280 & 0.0044 \\
math500 & MATH-500 & full & qwen+llama+tulu+dsmath+gemma+phi+ministral & 1.0 & 0.7236 & 0.0071 \\
math500 & MATH-500 & full & qwen+llama+tulu+dsmath+gemma+phi+ministral & 2.0 & 0.7202 & 0.0070 \\
gsm8k & GSM8K & full & qwen+llama+tulu+dsmath+gemma+phi+ministral & 0.0 & 0.9096 & 0.0016 \\
\bottomrule
\end{tabular}
\end{table*}

\begin{table*}[h!]
\centering
\scriptsize
\caption{K=50 distribution-level phase diagram. K=50 uses all cached samples once and serves as a robustness check for the main protocol.}
\label{tab:app-k50-phase}
\begin{tabular}{llllllll}
\toprule
Benchmark & Best acc. & Agg. & $\alpha$ & $\kappa$ & Majority & SC & Single-best \\
\midrule
MATH-500 & 0.8000 & Sum & 0.6316 & 7.0171 & 0.7200 & 0.7240 & 0.7980 \\
GSM8K & 0.9189 & Max & 1.0000 & 8.0000 & 0.9045 & 0.9052 & 0.9105 \\
GPQA & 0.4394 & Max & 0.6842 & 4.0481 & 0.3737 & 0.3030 & 0.4091 \\
MMLU-Pro & 0.6480 & Sum & 0.8947 & 0.0000 & 0.6280 & 0.5580 & 0.6180 \\
\bottomrule
\end{tabular}
\end{table*}

\begin{table*}[h!]
\centering
\scriptsize
\caption{Effective sample-budget scaling under K20R-style subsampling.}
\label{tab:app-keff}
\setlength{\tabcolsep}{2.5pt}
\renewcommand{\arraystretch}{0.95}

\resizebox{\textwidth}{!}{%
\begin{tabular}{lllllllllll}
\toprule
benchmark & benchmark\_label & pool\_role & pool\_id & K\_eff & R & method & accuracy\_mean & accuracy\_std & accuracy\_min & accuracy\_max \\
\midrule
math500 & MATH-500 & full & qwen+llama+tulu+dsmath+gemma+phi+ministral & 1.0000 & 50 & majority & 0.6561 & 0.0105 & 0.6340 & 0.6780 \\
math500 & MATH-500 & full & qwen+llama+tulu+dsmath+gemma+phi+ministral & 1.0000 & 50 & slm\_mux & 0.6561 & 0.0105 & 0.6340 & 0.6780 \\
math500 & MATH-500 & full & qwen+llama+tulu+dsmath+gemma+phi+ministral & 1.0000 & 50 & cc\_like & 0.6561 & 0.0105 & 0.6340 & 0.6780 \\
math500 & MATH-500 & full & qwen+llama+tulu+dsmath+gemma+phi+ministral & 1.0000 & 50 & single\_best & 0.7312 & 0.0111 & 0.7020 & 0.7540 \\
math500 & MATH-500 & full & qwen+llama+tulu+dsmath+gemma+phi+ministral & 1.0000 & 50 & best\_gen\_step3\_cell & 0.7315 & 0.0111 & 0.7020 & 0.7540 \\
math500 & MATH-500 & full & qwen+llama+tulu+dsmath+gemma+phi+ministral & 1.0000 & 50 & best\_sum\_step3\_cell & 0.7315 & 0.0111 & 0.7020 & 0.7540 \\
math500 & MATH-500 & full & qwen+llama+tulu+dsmath+gemma+phi+ministral & 1.0000 & 50 & best\_max\_step3\_cell & 0.7315 & 0.0111 & 0.7020 & 0.7540 \\
\bottomrule
\end{tabular}%
}
\end{table*}
\begin{table*}[h!]
\centering
\scriptsize
\caption{Method-correlation analysis. High correlation indicates that two methods draw on overlapping evidence.}
\label{tab:app-method-corr}
\begin{tabular}{lllll}
\toprule
benchmark & method\_a & method\_b & prediction\_agreement & correctness\_corr \\
\midrule
gpqa & single\_best\_full & majority\_full & 0.5955 & 0.4504 \\
gpqa & single\_best\_full & slm\_mux\_full & 0.5247 & 0.3339 \\
gpqa & single\_best\_full & cc\_like\_full & 0.5732 & 0.3978 \\
gpqa & single\_best\_full & best\_gen\_full & 0.7813 & 0.7442 \\
gpqa & single\_best\_full & oracle\_full & 0.4015 & 0.4282 \\
gpqa & majority\_full & slm\_mux\_full & 0.7869 & 0.7271 \\
gpqa & majority\_full & cc\_like\_full & 0.8929 & 0.8685 \\
\bottomrule
\end{tabular}
\end{table*}

\clearpage
\subsection{Atr/Ctr Model-Support Analysis}

This section asks which models the router listens to and whether that
support is helpful. Leave-one-out tables estimate the marginal effect of
removing each model. The Llama table separates pools that include Llama
from comparable Llama-free pools. The Atr/Ctr table reports the
counterfactual Ctr used in the main text. The accepted-support
decomposition separates support on correct routed answers from support on
wrong routed answers.

\begin{table*}[h!]
\centering
\scriptsize
\caption{Leave-one-out model contribution summary.}
\label{tab:app-leave-one-out}
\begin{tabular}{lllll}
\toprule
removed\_model & GPQA & GSM8K & MATH-500 & MMLU-Pro \\
\midrule
dsmath & -0.0015 & -0.0004 & 0.0002 & 0.0002 \\
gemma & 0.0076 & -0.0003 & -0.0002 & -0.0100 \\
llama & -0.0035 & 0.0036 & 0.0008 & 0.0054 \\
ministral & -0.0035 & -0.0008 & 0.0000 & -0.0006 \\
phi & -0.0066 & -0.0002 & -0.0006 & 0.0110 \\
qwen & 0.0485 & 0.0099 & 0.1118 & 0.0176 \\
tulu & -0.0056 & -0.0009 & 0.0000 & -0.0026 \\
\bottomrule
\end{tabular}
\end{table*}

\begin{table*}[h!]
\centering
\scriptsize
\caption{Llama inclusion analysis comparing the best pool containing Llama with the best pool excluding Llama.}
\label{tab:app-step4-e321-llama}
\begin{tabular}{lllrrr}
\toprule
Benchmark & Comparison & Pool & Size & Best gen. & $\Delta$ vs SLM \\
\midrule
GPQA & contains\_llama & qwen+llama+gemma+ministral & 4 & 0.4429 & 0.0212 \\
GPQA & no\_llama & qwen+gemma+phi & 3 & 0.4429 & 0.0152 \\
GSM8K & contains\_llama & qwen+llama+dsmath & 3 & 0.9327 & 0.0060 \\
GSM8K & no\_llama & qwen+dsmath & 2 & 0.9231 & 0.0016 \\
MATH-500 & contains\_llama & qwen+llama+dsmath+gemma & 4 & 0.7952 & 0.0558 \\
MATH-500 & no\_llama & qwen+dsmath+gemma & 3 & 0.7944 & 0.0546 \\
MMLU-Pro & contains\_llama & qwen+llama+dsmath+phi & 4 & 0.6526 & 0.0652 \\
MMLU-Pro & no\_llama & qwen+dsmath+gemma+phi & 4 & 0.6384 & 0.0594 \\
\bottomrule
\end{tabular}
\end{table*}

\begin{table*}[h!]
\centering
\scriptsize
\caption{Per-model Atr/Ctr diagnostics for the full pool and the selected curated pool in each benchmark. Atr is support assigned to the routed answer; Ctr is the leave-one-out counterfactual contribution defined in \S\ref{sec:method}.}
\label{tab:atr-ctr-by-model}
\begin{tabular}{lllrr}
\toprule
Benchmark & Setting & Model & Atr & Ctr \\
\midrule
MATH-500 & Full & qwen & 0.8283 & 0.1220 \\
MATH-500 & Full & llama & 0.5022 & 0.0014 \\
MATH-500 & Full & tulu & 0.4817 & 0.0002 \\
MATH-500 & Full & dsmath & 0.4392 & 0.0002 \\
MATH-500 & Full & gemma & 0.4951 & 0.0002 \\
MATH-500 & Full & phi & 0.4637 & -0.0004 \\
MATH-500 & Full & ministral & 0.5759 & 0.0012 \\
\bottomrule
\end{tabular}
\end{table*}

\begin{table*}[h!]
\centering
\scriptsize
\caption{Accepted-support decomposition. For each benchmark and pool role, we show the three models with highest Atr under best generalized routing. Balance is correct accepted support minus wrong accepted support.}
\label{tab:app-step4-e328-attribution}
\begin{tabular}{lllrrrrr}
\toprule
Benchmark & Pool & Model & Atr & Correct part & Wrong part & Balance & Precision \\
\midrule
GPQA & best\_abs & qwen & 0.6093 & 0.2862 & 0.3231 & -0.0368 & 0.4698 \\
GPQA & best\_abs & gemma & 0.5304 & 0.2480 & 0.2824 & -0.0343 & 0.4676 \\
GPQA & best\_abs & llama & 0.4682 & 0.2105 & 0.2578 & -0.0473 & 0.4496 \\
GPQA & full & qwen & 0.6139 & 0.2814 & 0.3326 & -0.0512 & 0.4583 \\
GPQA & full & gemma & 0.5169 & 0.2419 & 0.2750 & -0.0332 & 0.4679 \\
GPQA & full & phi & 0.4828 & 0.2148 & 0.2680 & -0.0532 & 0.4449 \\
GSM8K & best\_abs & qwen & 0.9179 & 0.8766 & 0.0413 & 0.8354 & 0.9550 \\
GSM8K & best\_abs & llama & 0.8482 & 0.8149 & 0.0332 & 0.7817 & 0.9608 \\
GSM8K & best\_abs & dsmath & 0.8121 & 0.7829 & 0.0292 & 0.7537 & 0.9640 \\
GSM8K & full & qwen & 0.9320 & 0.8757 & 0.0563 & 0.8194 & 0.9396 \\
GSM8K & full & tulu & 0.8722 & 0.8269 & 0.0453 & 0.7816 & 0.9481 \\
GSM8K & full & gemma & 0.8577 & 0.8069 & 0.0507 & 0.7562 & 0.9408 \\
MATH-500 & full & qwen & 0.8283 & 0.7189 & 0.1094 & 0.6094 & 0.8679 \\
MATH-500 & full & ministral & 0.5759 & 0.5335 & 0.0425 & 0.4910 & 0.9262 \\
MATH-500 & full & llama & 0.5022 & 0.4697 & 0.0325 & 0.4372 & 0.9353 \\
MMLU-Pro & full & qwen & 0.7273 & 0.5271 & 0.2002 & 0.3269 & 0.7247 \\
MMLU-Pro & full & gemma & 0.6430 & 0.4635 & 0.1795 & 0.2840 & 0.7209 \\
MMLU-Pro & full & llama & 0.5898 & 0.4385 & 0.1513 & 0.2871 & 0.7435 \\
\bottomrule
\end{tabular}
\end{table*}

\clearpage
\section{Supplementary Mechanism and Sensitivity Tables}

\subsection{Six-Model Exploratory Analyses}

These tables summarize a smaller six-model setting used to check the
same routing questions under a lighter cache. They report single-model
and baseline numbers, a fixed-prior generalized sweep, top pools from a
pool-composition scan, and sampling-strategy comparisons.

\begin{table*}[h!]
\centering
\scriptsize
\caption{Six-model exploratory baseline summary.}
\label{tab:app-step2-e100-baselines}
\begin{tabular}{llllrrr}
\toprule
Benchmark & Best single & Best single acc. & Majority & SLM-MUX & Cross-support & Single-Best \\
\midrule
MATH-500 & Qwen & 0.7660 & 0.7220 & 0.7200 & 0.7080 & 0.7680 \\
GSM8K & Qwen & 0.9022 & 0.8999 & 0.9158 & 0.9022 & 0.9037 \\
GPQA & Gemma & 0.4091 & 0.3636 & 0.3434 & 0.3535 & 0.3788 \\
\bottomrule
\end{tabular}
\end{table*}

\begin{table*}[h!]
\centering
\scriptsize
\caption{Six-model fixed-prior sweep summary.}
\label{tab:app-step2-e101}
\begin{tabular}{lllllllll}
\toprule
Benchmark & Majority & SC & Cross-support & Single-best & Best gen. & Best $\alpha$ & Best $\kappa$ & Agg. \\
\midrule
MATH-500 & 0.7060 & 0.7000 & 0.7040 & 0.7680 & 0.7760 & 0.3000 & 4.0000 & Max \\
GSM8K & 0.8999 & 0.9007 & 0.9014 & 0.9037 & 0.9105 & 1.0000 & 0.5000 & Max \\
GPQA & 0.3737 & 0.3535 & 0.3687 & 0.3990 & 0.4343 & 0.6000 & 4.0000 & Sum \\
\bottomrule
\end{tabular}
\end{table*}

\begin{table*}[h!]
\centering
\scriptsize
\caption{Six-model top pool-composition results by best generalized accuracy.}
\label{tab:app-step2-e102-top-pools}
\setlength{\tabcolsep}{3pt}
\renewcommand{\arraystretch}{0.95}

\resizebox{\textwidth}{!}{%
\begin{tabular}{llrlrrrrrl}
\toprule
Benchmark & Rank & Pool & Size & Best gen. & Single-best & $\Delta$ vs SB & $\Delta$ vs SLM & Oracle gap & Regime \\
\midrule
MATH-500 & 1 & qwen+phi & 2 & 0.7780 & 0.7680 & 0.0100 & 0.0280 & 0.0140 & low/moderate-alpha Sum coalition/Majority-like \\
MATH-500 & 2 & qwen+tulu+dsmath+phi+ministral & 5 & 0.7780 & 0.7680 & 0.0100 & 0.0660 & 0.0560 & high-kappa expert/single-best-like \\
MATH-500 & 3 & qwen+tulu+ministral & 3 & 0.7760 & 0.7680 & 0.0080 & 0.0340 & 0.0340 & high-kappa expert/single-best-like \\
GSM8K & 1 & qwen+tulu+dsmath & 3 & 0.9219 & 0.9037 & 0.0182 & 0.0045 & 0.0569 & low/moderate-alpha Sum coalition/Majority-like \\
GSM8K & 2 & qwen+dsmath+phi & 3 & 0.9189 & 0.9037 & 0.0152 & 0.0099 & 0.0584 & high-kappa expert/single-best-like \\
GSM8K & 3 & qwen+dsmath & 2 & 0.9174 & 0.9037 & 0.0136 & 0.0038 & 0.0478 & low/moderate-alpha Sum coalition/Majority-like \\
GPQA & 1 & qwen+gemma+phi & 3 & 0.4495 & 0.3990 & 0.0505 & 0.0152 & 0.2525 & low/moderate-alpha Sum coalition/Majority-like \\
GPQA & 2 & qwen+dsmath+gemma+phi & 4 & 0.4495 & 0.3990 & 0.0505 & 0.0758 & 0.2828 & low/moderate-alpha Sum coalition/Majority-like \\
GPQA & 3 & qwen+gemma & 2 & 0.4394 & 0.3990 & 0.0404 & 0.0202 & 0.1667 & low/moderate-alpha Sum coalition/Majority-like \\
\bottomrule
\end{tabular}%
}
\end{table*}
\begin{table*}[h!]
\centering
\scriptsize
\caption{Sampling-strategy comparison. K20R10 preserves replicate variability while using the larger cache.}
\label{tab:app-step2-e103}
\begin{tabular}{llrrrrrrrl}
\toprule
Benchmark & Pool & Strategy & Majority & SC & Cross-support & Single-best & Best gen. & Std & Modal best params \\
\midrule
GPQA & six-model & K20 once & 0.3932 & 0.3422 & 0.3730 & 0.4020 & 0.4444 & 0.0087 & Sum a=0.9 k=2 (2/20) \\
GPQA & six-model & K20R10 mean & 0.3821 & 0.3172 & 0.3750 & 0.4078 & 0.4467 & 0.0081 & Max a=0.9 k=2 (6/20) \\
GPQA & six-model & K=50 full & 0.3838 & 0.3081 & 0.3788 & 0.4091 & 0.4495 & 0.0000 & Max a=0.9 k=2 (1/1) \\
GPQA & seven-model & K20 once & 0.3929 & 0.3409 & 0.3727 & 0.4020 & 0.4394 & 0.0118 & Sum a=0.0 k=8 (3/20) \\
GPQA & seven-model & K20R10 mean & 0.3798 & 0.3096 & 0.3674 & 0.4078 & 0.4379 & 0.0058 & Max a=0.9 k=2 (4/20) \\
GPQA & seven-model & K=50 full & 0.3737 & 0.3030 & 0.3535 & 0.4091 & 0.4343 & 0.0000 & Sum a=0.2 k=8 (1/1) \\
GSM8K & six-model & K20 once & 0.8989 & 0.8983 & 0.9010 & 0.9080 & 0.9135 & 0.0018 & Max a=1.0 k=4 (13/20) \\
\bottomrule
\end{tabular}
\end{table*}

\clearpage
\subsection{Seven-Model Protocol Checks}

These tables record seven-model protocol checks: comparable baselines,
replicate-aware generalized sweeps, and pool-subset scans under the
K20R10 protocol.

\begin{table*}[h!]
\centering
\scriptsize
\caption{Seven-model K20R10 baseline summary. Standard deviations are across the 10 subsampling replicates.}
\label{tab:app-step3-e200-baselines}
\begin{tabular}{lllllll}
\toprule
Benchmark & Best single & Best single & Majority & SLM-MUX & Cross slice & Oracle \\
\midrule
MATH-500 & qwen & 0.7924 $\pm$ 0.0048 & 0.7246 $\pm$ 0.0073 & 0.7242 $\pm$ 0.0053 & 0.7302 $\pm$ 0.0040 & 0.8536 $\pm$ 0.0067 \\
GSM8K & qwen & 0.9089 $\pm$ 0.0024 & 0.9032 $\pm$ 0.0018 & 0.9031 $\pm$ 0.0018 & 0.9050 $\pm$ 0.0015 & 0.9760 $\pm$ 0.0011 \\
GPQA & qwen & 0.4015 $\pm$ 0.0162 & 0.3864 $\pm$ 0.0069 & 0.3369 $\pm$ 0.0155 & 0.3727 $\pm$ 0.0087 & 0.7854 $\pm$ 0.0116 \\
MMLU-Pro & qwen & 0.6158 $\pm$ 0.0058 & 0.6282 $\pm$ 0.0055 & 0.5738 $\pm$ 0.0085 & 0.6130 $\pm$ 0.0045 & 0.8420 $\pm$ 0.0042 \\
\bottomrule
\end{tabular}
\end{table*}

\begin{table*}[h!]
\centering
\scriptsize
\caption{Seven-model K20R10 fixed-prior sweep summary.}
\label{tab:app-step3-e201}
\begin{tabular}{lllllllll}
\toprule
Benchmark & Majority & SC & Cross-support & Single-best & Best gen. & Best $\alpha$ & Best $\kappa$ & Agg. \\
\midrule
MATH-500 & 0.7246 & 0.7242 & 0.7302 & 0.7924 & 0.7938 & 0.6842 & 7.0171 & Sum \\
GSM8K & 0.9032 & 0.9031 & 0.9050 & 0.9089 & 0.9166 & 1.0000 & 7.0171 & Max \\
GPQA & 0.3864 & 0.3369 & 0.3727 & 0.4015 & 0.4328 & 0.3684 & 8.0000 & Max \\
MMLU-Pro & 0.6282 & 0.5738 & 0.6130 & 0.6158 & 0.6402 & 0.1053 & 5.3617 & Max \\
\bottomrule
\end{tabular}
\end{table*}

\begin{table*}[h!]
\centering
\scriptsize
\caption{Seven-model top pool-composition results, with five best pools per benchmark by generalized accuracy.}
\label{tab:app-step3-e202-top-pools}
\begin{tabular}{lp{0.32\linewidth}rrrrrl}
\toprule
Benchmark & Pool & Size & Best gen. & Single-best & SC & Oracle & Agg. \\
\midrule
MATH-500 & qwen+llama+tulu+gemma & 4 & 0.7952 & 0.7924 & 0.7452 & 0.8366 & Max \\
MATH-500 & qwen+llama+dsmath+gemma & 4 & 0.7952 & 0.7924 & 0.7394 & 0.8380 & Max \\
MATH-500 & qwen+llama+gemma & 3 & 0.7950 & 0.7924 & 0.7504 & 0.8294 & Sum \\
MATH-500 & qwen+llama+tulu+dsmath+gemma & 5 & 0.7948 & 0.7924 & 0.7360 & 0.8434 & Sum \\
MATH-500 & qwen+llama & 2 & 0.7944 & 0.7924 & 0.7860 & 0.8174 & Sum \\
GSM8K & qwen+llama+dsmath & 3 & 0.9327 & 0.9089 & 0.9267 & 0.9660 & Sum \\
GSM8K & qwen+llama+tulu+dsmath & 4 & 0.9281 & 0.9089 & 0.9252 & 0.9678 & Max \\
GSM8K & qwen+llama & 2 & 0.9238 & 0.9089 & 0.9213 & 0.9479 & Sum \\
GSM8K & qwen+dsmath & 2 & 0.9231 & 0.9089 & 0.9215 & 0.9569 & Sum \\
GSM8K & qwen+tulu+dsmath & 3 & 0.9230 & 0.9089 & 0.9227 & 0.9635 & Max \\
GPQA & qwen+gemma+phi & 3 & 0.4429 & 0.4015 & 0.4278 & 0.6348 & Max \\
GPQA & qwen+llama+gemma+ministral & 4 & 0.4429 & 0.4015 & 0.4217 & 0.6788 & Sum \\
GPQA & qwen+llama+gemma & 3 & 0.4424 & 0.4015 & 0.4247 & 0.6343 & Sum \\
GPQA & qwen+llama+gemma+phi & 4 & 0.4409 & 0.4015 & 0.4162 & 0.6904 & Max \\
GPQA & qwen+dsmath+gemma+phi & 4 & 0.4409 & 0.4015 & 0.3591 & 0.6970 & Max \\
MMLU-Pro & qwen+llama+dsmath+phi & 4 & 0.6526 & 0.6158 & 0.5874 & 0.7846 & Sum \\
MMLU-Pro & qwen+llama+phi & 3 & 0.6506 & 0.6158 & 0.6192 & 0.7528 & Sum \\
MMLU-Pro & qwen+llama+tulu+dsmath+phi+ministral & 6 & 0.6502 & 0.6158 & 0.5660 & 0.8274 & Sum \\
MMLU-Pro & qwen+llama+tulu+dsmath+phi & 5 & 0.6486 & 0.6158 & 0.5672 & 0.8080 & Sum \\
MMLU-Pro & qwen+llama+tulu+phi+ministral & 5 & 0.6478 & 0.6158 & 0.5804 & 0.7992 & Sum \\
\bottomrule
\end{tabular}
\end{table*}

\clearpage
\subsection{Additional Mechanism Tables}

The following tables probe mechanism questions that complement the main
analysis: whether cross-support is a question-specific signal, how it
behaves under placebo tests, how sample budget affects routing signals,
and where oracle coverage fails to become routable support.

\begin{table*}[h!]
\centering
\small
\caption{Row-level correctness prediction quality for isolated self-consistency $c$, cross-support $s$, and $c{+}s$.}
\label{tab:signal-metrics}
\begin{tabular}{lrrrrrr}
\toprule
Benchmark & Rows & AUC$(c)$ & AUC$(s)$ & AUC$(c{+}s)$ & $\rho(c,y)$ & $\rho(s,y)$ \\
\midrule
MATH-500 & 29780 & 0.820 & 0.904 & 0.901 & 0.574 & 0.706 \\
GSM8K & 79140 & 0.754 & 0.945 & 0.936 & 0.438 & 0.581 \\
GPQA & 11280 & 0.525 & 0.600 & 0.574 & 0.043 & 0.164 \\
GPQA ($K{=}3$) & 11088 & 0.531 & 0.605 & 0.584 & 0.055 & 0.172 \\
\bottomrule
\end{tabular}
\end{table*}

\begin{table*}[h!]
\centering
\scriptsize
\caption{Placebo tests that corrupt cross-support while leaving self-consistency fixed.}
\label{tab:placebo}
\begin{tabular}{lrrrrr}
\toprule
Setting & SLM & Cross real & Cross q-shuf & Cross m-shuf & $\Delta$ real \\
\midrule
GPQA / 6-model / K=5 & 0.343 & 0.354 & 0.345 & 0.320 & 0.010 \\
GPQA / 6-model / K=3 & 0.332 & 0.363 & 0.341 & 0.317 & 0.031 \\
GPQA / qwen+phi+gemma / K=5 & 0.394 & 0.429 & 0.404 & 0.406 & 0.035 \\
MATH-500 / 6-model / K=5 & 0.720 & 0.708 & 0.704 & 0.692 & -0.012 \\
GSM8K / 6-model / K=5 & 0.916 & 0.902 & 0.908 & 0.892 & -0.014 \\
\bottomrule
\end{tabular}
\end{table*}

\begin{table*}[h!]
\centering
\scriptsize
\caption{Best score form in each setting from the score-form ablation.}
\label{tab:score-form}
\begin{tabular}{llrr}
\toprule
Setting & Best variant & Acc. & $\Delta$ vs SLM \\
\midrule
GPQA / 6-model / K=3 & s\_only & 0.388 & 0.056 \\
GPQA / 6-model / K=5 & s\_only & 0.379 & 0.035 \\
GPQA / qwen+phi+gemma / K=5 & rank\_sum & 0.439 & 0.045 \\
MATH-500 / 6-model / K=5 & c\_times\_s & 0.722 & 0.002 \\
GSM8K / 6-model / K=5 & c\_only & 0.916 & 0.000 \\
\bottomrule
\end{tabular}
\end{table*}

\begin{table*}[h!]
\centering
\scriptsize
\caption{Best support definition by $s$-only accuracy.}
\label{tab:support-def}
\begin{tabular}{llrr}
\toprule
Setting & Support definition & $s$-only & $\Delta$ vs SLM \\
\midrule
GPQA / 6-model / K=3 & top2\_strong & 0.396 & 0.064 \\
GPQA / 6-model / K=5 & top2\_strong & 0.419 & 0.076 \\
GPQA / qwen+phi+gemma / K=5 & majority\_fraction & 0.414 & 0.020 \\
\bottomrule
\end{tabular}
\end{table*}

\begin{table*}[h!]
\centering
\scriptsize
\caption{Best normalisation / calibration configuration in each setting.}
\label{tab:calibration}
\begin{tabular}{llrr}
\toprule
Setting & Best configuration & Acc. & $\Delta$ vs SLM \\
\midrule
GPQA / 6-model / K=5 & s\_only (rank) & 0.384 & 0.040 \\
GPQA / qwen+phi+gemma / K=5 & c+s (zscore,rank) & 0.449 & 0.056 \\
MATH-500 / 6-model / K=5 & c+s (minmax,minmax) & 0.720 & 0.000 \\
\bottomrule
\end{tabular}
\end{table*}

\begin{table*}[h!]
\centering
\scriptsize
\caption{How the routing signals change with the sample budget $K$.}
\label{tab:k-signal}
\begin{tabular}{lrrrrrr}
\toprule
Benchmark & $K$ & AUC$(c)$ & AUC$(s)$ & AUC$(c{+}s)$ & Entropy & Stable-wrong SLM \\
\midrule
GPQA & 1 & 0.500 & 0.599 & 0.599 & 1.218 & 64.6 \\
GPQA & 2 & 0.522 & 0.608 & 0.582 & 1.401 & 64.9 \\
GPQA & 3 & 0.531 & 0.605 & 0.584 & 1.460 & 66.8 \\
GPQA & 4 & 0.535 & 0.602 & 0.583 & 1.485 & 66.7 \\
GPQA & 5 & 0.525 & 0.600 & 0.574 & 1.506 & 65.7 \\
MATH-500 & 1 & 0.500 & 0.944 & 0.944 & 1.272 & 27.6 \\
MATH-500 & 2 & 0.784 & 0.954 & 0.948 & 1.627 & 25.9 \\
MATH-500 & 3 & 0.837 & 0.953 & 0.949 & 1.803 & 26.5 \\
MATH-500 & 4 & 0.854 & 0.952 & 0.949 & 1.926 & 24.0 \\
MATH-500 & 5 & 0.861 & 0.951 & 0.948 & 2.011 & 24.2 \\
GSM8K & 1 & 0.500 & 0.939 & 0.939 & 0.417 & 11.4 \\
GSM8K & 2 & 0.712 & 0.950 & 0.944 & 0.486 & 9.4 \\
GSM8K & 3 & 0.735 & 0.947 & 0.940 & 0.515 & 9.2 \\
GSM8K & 4 & 0.748 & 0.946 & 0.938 & 0.535 & 8.6 \\
GSM8K & 5 & 0.754 & 0.945 & 0.936 & 0.548 & 8.4 \\
\bottomrule
\end{tabular}
\end{table*}

\begin{table*}[h!]
\centering
\scriptsize
\caption{Accuracy scaling with the per-model sample budget $K$ on the full 6-model pool.}
\label{tab:k-scaling}
\begin{tabular}{lrrrrrr}
\toprule
Benchmark & $K$ & SLM & Cross & $\Delta$ & Fix & Break \\
\midrule
MATH-500 & 1 & 0.724 & 0.703 & -0.022 & 116 & 224 \\
MATH-500 & 2 & 0.708 & 0.694 & -0.014 & 94 & 166 \\
MATH-500 & 3 & 0.711 & 0.703 & -0.009 & 81 & 124 \\
MATH-500 & 4 & 0.713 & 0.708 & -0.005 & 76 & 101 \\
MATH-500 & 5 & 0.720 & 0.708 & -0.012 & 60 & 120 \\
GSM8K & 1 & 0.886 & 0.897 & 0.011 & 405 & 254 \\
GSM8K & 2 & 0.902 & 0.899 & -0.003 & 212 & 251 \\
GSM8K & 3 & 0.908 & 0.899 & -0.009 & 149 & 268 \\
GSM8K & 4 & 0.912 & 0.900 & -0.012 & 133 & 295 \\
GSM8K & 5 & 0.916 & 0.902 & -0.014 & 130 & 310 \\
GPQA & 1 & 0.354 & 0.364 & 0.010 & 201 & 181 \\
GPQA & 2 & 0.350 & 0.366 & 0.016 & 172 & 140 \\
GPQA & 3 & 0.332 & 0.363 & 0.031 & 165 & 104 \\
GPQA & 4 & 0.334 & 0.352 & 0.018 & 143 & 107 \\
GPQA & 5 & 0.343 & 0.354 & 0.010 & 110 & 90 \\
\bottomrule
\end{tabular}
\end{table*}

\begin{table*}[h!]
\centering
\scriptsize
\caption{Oracle-gap decomposition of routing outcomes.}
\label{tab:oracle-decomp}
\begin{tabular}{lrrrrrr}
\toprule
Setting & Routed correct & No oracle & Stable-but-wrong & No support & Invalid & Other \\
\midrule
GPQA / 6-model / K=5 & 34.3 & 24.2 & 31.8 & 4.5 & 0.0 & 5.1 \\
GPQA / qwen+phi+gemma / K=5 & 39.4 & 34.8 & 16.7 & 5.1 & 0.0 & 4.0 \\
MATH-500 / 6-model / K=5 & 72.0 & 16.6 & 9.8 & 0.6 & 0.0 & 1.0 \\
GSM8K / 6-model / K=5 & 91.6 & 2.9 & 2.4 & 2.2 & 0.0 & 1.0 \\
\bottomrule
\end{tabular}
\end{table*}

\begin{table*}[h!]
\centering
\scriptsize
\caption{Top GPQA model pairs by pairwise oracle. All top pairs contain Gemma.}
\label{tab:pairwise}
\begin{tabular}{lrrrr}
\toprule
Pair & Pair oracle & Inc. gain & Error overlap & Wrong-answer agr. \\
\midrule
qwen + gemma & 0.566 & 0.167 & 0.434 & 0.477 \\
tulu + gemma & 0.561 & 0.162 & 0.439 & 0.471 \\
gemma + phi & 0.556 & 0.157 & 0.444 & 0.409 \\
qwen + tulu & 0.545 & 0.167 & 0.455 & 0.544 \\
qwen + ministral & 0.535 & 0.157 & 0.465 & 0.533 \\
\bottomrule
\end{tabular}
\end{table*}

\begin{table*}[h!]
\centering
\scriptsize
\caption{Question-slice analysis on GPQA by model disagreement.}
\label{tab:question-slices}
\begin{tabular}{lrrrrrr}
\toprule
Setting & Bin & Views & SLM & Cross & $s$-only & $\Delta$ Cross-SLM \\
\midrule
GPQA / 6-model / K=3 & q1 & 426 & 0.434 & 0.467 & 0.460 & 0.033 \\
GPQA / 6-model / K=3 & q2 & 323 & 0.359 & 0.399 & 0.424 & 0.040 \\
GPQA / 6-model / K=3 & q3 & 317 & 0.341 & 0.404 & 0.423 & 0.063 \\
GPQA / 6-model / K=3 & q4 & 914 & 0.271 & 0.287 & 0.329 & 0.015 \\
GPQA / 6-model / K=5 & q1 & 430 & 0.442 & 0.465 & 0.442 & 0.023 \\
GPQA / 6-model / K=5 & q2 & 380 & 0.368 & 0.474 & 0.500 & 0.105 \\
GPQA / 6-model / K=5 & q3 & 230 & 0.391 & 0.391 & 0.391 & 0.000 \\
GPQA / 6-model / K=5 & q4 & 940 & 0.277 & 0.245 & 0.298 & -0.032 \\
\bottomrule
\end{tabular}
\end{table*}

\begin{table}[h!]
\centering
\small
\caption{OLS coefficients for predicting $\Delta = \text{Cross} - \text{SLM}$ from subset-level pool statistics.}
\label{tab:regime-regression}
\begin{tabular}{lrrrrr}
\toprule
Benchmark & $\beta_{\text{dom}}$ & $\beta_{\text{oracle}}$ & $\beta_{\text{size}}$ & $R^2$ & $n$ \\
\midrule
MATH-500 & 0.015 & 0.223 & -0.005 & 0.669 & 57 \\
GSM8K & 0.125 & 0.270 & -0.004 & 0.655 & 57 \\
GPQA & 0.072 & 0.212 & -0.013 & 0.106 & 57 \\
\bottomrule
\end{tabular}
\end{table}

\begin{table*}[h!]
\centering
\scriptsize
\caption{GPQA contamination analysis anchored at the curated core pool ``qwen\_phi\_gemma``.}
\label{tab:contamination}
\begin{tabular}{lrrrrr}
\toprule
Pool & $n$ & Cross & $\Delta$ vs core & Oracle & $\Delta$ oracle \\
\midrule
qwen\_phi\_gemma & 3 & 0.439 & 0.000 & 0.652 & 0.000 \\
qwen\_tulu\_phi\_gemma & 4 & 0.384 & -0.056 & 0.712 & 0.061 \\
qwen\_phi\_ministral\_gemma & 4 & 0.429 & -0.010 & 0.702 & 0.051 \\
qwen\_phi\_gemma\_dsmath & 4 & 0.374 & -0.066 & 0.682 & 0.030 \\
qwen\_tulu\_phi\_gemma & 4 & 0.384 & -0.056 & 0.712 & 0.061 \\
qwen\_tulu\_phi\_ministral\_gemma & 5 & 0.364 & -0.076 & 0.747 & 0.096 \\
qwen\_tulu\_phi\_ministral\_gemma\_dsmath & 6 & 0.359 & -0.081 & 0.758 & 0.106 \\
qwen\_gemma & 2 & 0.434 & -0.005 & 0.566 & -0.086 \\
phi\_gemma & 2 & 0.419 & -0.020 & 0.556 & -0.096 \\
\bottomrule
\end{tabular}
\end{table*}

\begin{table*}[h!]
\centering
\scriptsize
\caption{Question-level paired bootstrap confidence intervals and permutation tests.}
\label{tab:significance}
\begin{tabular}{lrrrr}
\toprule
Comparison & $N$ & $\Delta$ & 95\% CI & $p$ \\
\midrule
GPQA / 6-model / K=5  : cross-support vs slm\_mux & 198 & 0.010 & [-0.035, 0.056] & 0.822 \\
GPQA / 6-model / K=3  : cross-support vs slm\_mux & 198 & 0.031 & [-0.008, 0.068] & 0.115 \\
GPQA / qwen+phi+gemma / K=5: cross-support vs slm\_mux & 198 & 0.005 & [-0.030, 0.045] & 1.000 \\
GPQA / qwen+phi+gemma / K=5: cross-support vs single\_best\_sc & 198 & 0.040 & [-0.020, 0.101] & 0.249 \\
MATH-500 / 6-model / K=5: cross-support vs slm\_mux & 500 & -0.012 & [-0.028, 0.004] & 0.238 \\
GSM8K   / 6-model / K=5: cross-support vs slm\_mux & 1319 & -0.014 & [-0.024, -0.004] & 0.011 \\
\bottomrule
\end{tabular}
\end{table*}

\begin{table*}[h!]
\centering
\scriptsize
\caption{Tie-break robustness under model-order and validation-accuracy protocols.}
\label{tab:tiebreak}
\begin{tabular}{lllrrr}
\toprule
Setting & Tie-break & Acc source & SLM & Cross & $\Delta$ \\
\midrule
GPQA / 6-model / K=5 & model\_order & ignore & 0.343 & 0.354 & 0.010 \\
GPQA / 6-model / K=5 & validation\_accuracy & auto & 0.354 & 0.359 & 0.005 \\
GPQA / qwen+phi+gemma / K=5 & model\_order & ignore & 0.394 & 0.429 & 0.035 \\
GPQA / qwen+phi+gemma / K=5 & validation\_accuracy & auto & 0.434 & 0.439 & 0.005 \\
MATH-500 / 6-model / K=5 & model\_order & ignore & 0.720 & 0.708 & -0.012 \\
MATH-500 / 6-model / K=5 & validation\_accuracy & auto & 0.722 & 0.710 & -0.012 \\
\bottomrule
\end{tabular}
\end{table*}

\begin{table}[h!]
\centering
\small
\caption{Held-out GPQA pool selection over 20 random 50/50 splits.}
\label{tab:heldout-selection}
\begin{tabular}{lr}
\toprule
Method & Held-out mean \\
\midrule
Selected small-pool cross-support & 0.415 \\
Full-pool cross-support & 0.364 \\
Full-pool SLM & 0.345 \\
Full-pool Single-Best & 0.388 \\
\bottomrule
\end{tabular}
\end{table}
\noindent\textit{Selection frequencies:} qwen\_gemma: 9, qwen\_phi\_gemma: 6, qwen\_phi\_ministral\_gemma: 4, phi\_gemma: 1.

\begin{table*}[h!]
\centering
\scriptsize
\caption{Seed robustness over 20 offline subsampling seeds.}
\label{tab:seed-robustness}
\begin{tabular}{lrrrr}
\toprule
Setting & SLM & Cross & $\Delta$ & Wins / ties / losses \\
\midrule
GPQA / 6-model / K=3 & 0.340 $\pm$ 0.006 & 0.360 $\pm$ 0.006 & 0.020 $\pm$ 0.008 & 19/0/1 \\
GPQA / 6-model / K=5 & 0.343 $\pm$ 0.000 & 0.354 $\pm$ 0.000 & 0.010 $\pm$ 0.000 & 20/0/0 \\
GPQA / qwen+phi+gemma / K=5 & 0.394 $\pm$ 0.000 & 0.429 $\pm$ 0.000 & 0.035 $\pm$ 0.000 & 20/0/0 \\
MATH-500 / 6-model / K=5 & 0.720 $\pm$ 0.000 & 0.708 $\pm$ 0.000 & -0.012 $\pm$ 0.000 & 0/0/20 \\
\bottomrule
\end{tabular}
\end{table*}

\begin{table*}[h!]
\centering
\scriptsize
\caption{Budget-fair comparison on GPQA. Curated pools with deeper sampling dominate the full 6-model reference.}
\label{tab:budget-fairness}
\begin{tabular}{lrrrrr}
\toprule
Configuration & Budget & SLM & Cross & Single-Best & Oracle \\
\midrule
6 models $\times$ K=1 (budget=6) & 6 & 0.354 & 0.364 & 0.353 & 0.766 \\
4 models $\times$ K=2 (budget=8)  qwen+phi+ministral+gemma & 8 & 0.350 & 0.372 & 0.360 & 0.684 \\
3 models $\times$ K=3 (budget=9)  qwen+phi+gemma & 9 & 0.376 & 0.400 & 0.371 & 0.641 \\
2 models $\times$ K=5 (budget=10) qwen+gemma & 10 & 0.409 & 0.414 & 0.379 & 0.566 \\
6 models $\times$ K=5 (budget=30) full pool reference & 30 & 0.343 & 0.354 & 0.379 & 0.758 \\
\bottomrule
\end{tabular}
\end{table*}

\begin{table}[h!]
\centering
\small
\caption{With- vs without-validation comparison. Validation-selected single-best is the strongest deployable baseline.}
\label{tab:validation}
\begin{tabular}{lrrrrrr}
\toprule
Benchmark & Order SB & Val SB & SLM & Cross & $s$-only & Composite \\
\midrule
MATH-500 & 0.768 & 0.768 & 0.720 & 0.708 & 0.712 & 0.768 \\
GSM8K & 0.904 & 0.904 & 0.916 & 0.902 & 0.899 & 0.902 \\
GPQA & 0.379 & 0.399 & 0.343 & 0.354 & 0.379 & 0.354 \\
\bottomrule
\end{tabular}
\end{table}

\begin{table*}[h!]
\centering
\scriptsize
\caption{Held-out GPQA pool selection across four selection rules.}
\label{tab:pool-selection}
\begin{tabular}{lrrrr}
\toprule
Rule & Mean test cross-support & Std & Mean test $s$-only & Most selected pool \\
\midrule
max\_calib\_cross & 0.417 & 0.022 & 0.392 & qwen\_phi\_gemma \\
max\_calib\_s\_only & 0.408 & 0.028 & 0.381 & qwen\_gemma \\
min\_pool\_size & 0.420 & 0.022 & 0.393 & qwen\_gemma \\
smallest\_top\_cross & 0.414 & 0.023 & 0.394 & qwen\_gemma \\
\bottomrule
\end{tabular}
\end{table*}

\begin{table}[h!]
\centering
\small
\caption{Case taxonomy for the GPQA qualitative analysis.}
\label{tab:case-taxonomy}
\begin{tabular}{lrr}
\toprule
Category & Rows & Fraction \\
\midrule
Both wrong, oracle hit & 710 & 61.2 \\
Cross-support fixes SLM & 110 & 9.5 \\
Cross-support breaks SLM & 90 & 7.8 \\
Contamination & 250 & 21.6 \\
\bottomrule
\end{tabular}
\end{table}

\clearpage
\subsection{Additional Generalized-Family Checks}

These generalized-family tables provide implementation checks,
canonical-point recovery, K interactions, validation-aware variants,
region summaries, significance checks, and budget-fair comparisons.

\begin{table*}[h!]
\centering
\scriptsize
\caption{Implementation-level recovery of the canonical baselines as
limits inside the generalized family. All limits are recovered
exactly modulo the deterministic $(W, \text{pooled}, \text{lex})$
tie-break specified for the family; the small MATH-500 majority gap is
from a handful of questions on which one model has fewer valid extracted
answers than the others.}
\label{tab:generalized-sanity}
\begin{tabular}{lrrrrr}
\toprule
Benchmark / $K$ & majority $\delta$ & top-coal $\delta$ & SLM-MUX$_\text{ans}$ $\delta$ & cross slice $\delta$ & single-best $\delta$ \\
\midrule
GPQA / $K{=}5$ / 6m       & $+0.000$ & --       & $+0.010$ & --       & $+0.000$ \\
MATH-500 / $K{=}5$ / 6m   & $-0.004$ & --       & $-0.020$ & --       & $+0.000$ \\
GSM8K / $K{=}5$ / 6m      & $+0.000$ & --       & $-0.015$ & --       & $+0.000$ \\
\bottomrule
\end{tabular}
\end{table*}

\begin{table*}[h!]
\centering
\scriptsize
\caption{Generalized two-axis family on the full 6-model and curated
\gpqa pools. All canonical points are evaluated on the same
cached generations as the rest of the paper. $\alpha$ controls
majority-vs-top-answer aggregation; $\kappa$ controls uniform-vs-expert
weighting (calibration-half acc$_i$, held-out evaluation for
$\kappa{>}0$); Agg selects coalition (Sum) vs strongest-supporter (Max)
aggregation. Best generalized point in bold per row.}
\label{tab:generalized-canonical}
\begin{tabular}{lrrrrrrrr}
\toprule
Setting & majority & top-coal & SLM-MUX$_\text{ans}$ & cross slice & best Sum & best Max & best $(\alpha, \kappa, \text{Agg})$ & best acc \\
 & ($\alpha{=}0$) & ($\alpha{=}1$, Sum) & ($\alpha{=}1$, Max) & ($\alpha{=}1{-}\tfrac{1}{N-1}$) & ($\kappa{=}0$) & ($\kappa{=}0$) & & \\
\midrule
GPQA / 6m / $K{=}3$ & 0.384 & 0.372 & 0.364 & 0.365 & 0.384 & 0.384 & ($0.0,\ 0,\ \text{Sum}$) & \textbf{0.384} \\
GPQA / 6m / $K{=}5$ & 0.374 & 0.369 & 0.354 & 0.369 & 0.379 & 0.374 & ($1.0,\ 4,\ \text{Max}$) & \textbf{0.394} \\
GPQA / qwen+phi+gemma / $K{=}3$ & 0.410 & 0.394 & 0.405 & 0.404 & 0.410 & 0.410 & ($0.0,\ 0,\ \text{Sum}$) & \textbf{0.410} \\
GPQA / qwen+phi+gemma / $K{=}5$ & 0.419 & 0.424 & 0.434 & 0.429 & 0.450 & 0.439 & ($0.6,\ 0,\ \text{Sum}$) & \textbf{0.450} \\
GPQA / qwen+gemma / $K{=}5$ & 0.429 & 0.419 & 0.419 & 0.429 & 0.429 & 0.424 & ($0.0,\ 0,\ \text{Sum}$) & \textbf{0.429} \\
GPQA / phi+gemma / $K{=}5$ & 0.404 & 0.419 & 0.419 & 0.404 & 0.404 & 0.419 & ($0.1,\ 2,\ \text{Sum}$) & \textbf{0.444} \\
MATH-500 / 6m / $K{=}5$ & 0.706 & 0.702 & 0.700 & 0.704 & 0.708 & 0.712 & ($0.1,\ 4,\ \text{Sum}$) & \textbf{0.800} \\
GSM8K / 6m / $K{=}5$ & 0.900 & 0.897 & 0.901 & 0.901 & 0.900 & 0.901 & ($1.0,\ 0.5,\ \text{Max}$) & \textbf{0.909} \\
\bottomrule
\end{tabular}
\end{table*}

\begin{table*}[h!]
\centering
\scriptsize
\caption{K interaction on \gpqa for the generalized family. Best
validation-free $\kappa{=}0$ generalized point reported per $K$, alongside
the canonical points. ``--'' marks aggregations not swept at
$K{=}2$ and $K{=}4$. The full pool peaks at
$K{=}3$ in the $\kappa{=}0$ region; the curated pool is monotonic in $K$.}
\label{tab:generalized-k-interaction}
\begin{tabular}{lrrrrrcr}
\toprule
Pool & $K$ & majority & SLM-MUX$_\text{ans}$ & cross slice & single-best (val) & best gen $(\alpha, \kappa, \text{Agg})$ & best acc \\
\midrule
full 6m         & 1 & 0.370 & 0.370 & 0.370 & 0.338 & ($0.0,0,\text{Sum}$) & 0.370 \\
full 6m         & 2 & 0.372 & 0.372 & 0.372 & 0.300 & --                    & -- \\
full 6m         & 3 & 0.384 & 0.364 & 0.365 & 0.342 & ($0.0,0,\text{Sum}$) & \textbf{0.384} \\
full 6m         & 4 & 0.365 & 0.349 & 0.358 & 0.369 & --                    & -- \\
full 6m         & 5 & 0.374 & 0.354 & 0.369 & 0.364 & ($1.0,4,\text{Max}$) & 0.394 \\
qwen+phi+gemma  & 1 & 0.375 & 0.375 & 0.375 & 0.334 & ($0.0,0,\text{Sum}$) & 0.375 \\
qwen+phi+gemma  & 2 & 0.389 & 0.385 & 0.385 & 0.300 & --                    & -- \\
qwen+phi+gemma  & 3 & 0.410 & 0.405 & 0.404 & 0.337 & ($0.0,0,\text{Sum}$) & 0.410 \\
qwen+phi+gemma  & 4 & 0.415 & 0.410 & 0.411 & 0.369 & --                    & -- \\
qwen+phi+gemma  & 5 & 0.419 & 0.434 & 0.429 & 0.364 & ($0.6,0,\text{Sum}$) & \textbf{0.450} \\
\bottomrule
\end{tabular}
\end{table*}

\begin{table*}[h!]
\centering
\scriptsize
\caption{Validation-aware evaluation of the generalized
family. Per setting we draw a fresh 50/50 random split over
questions for each of 20 seeds, fit per-model calibration accuracies
$\text{acc}_i$ on one half, and evaluate every $(\alpha, \kappa,
\text{Agg})$ cell on the held-out half. All numbers are held-out
accuracy averaged across the 20 seeds. ``Single-best (val)'' is the
$\kappa{=}\infty$ limit.}
\label{tab:generalized-validation-aware}
\begin{tabular}{lrrrrrr}
\toprule
Setting & majority & SLM-MUX$_\text{ans}$ & cross slice & single-best (val) & best Sum & best Max \\
\midrule
GPQA / 6m / $K{=}3$           & 0.384 & 0.362 & 0.364 & 0.360 & \textbf{0.391}@($0.0,1.0$) & 0.391@($0.0,1.0$) \\
GPQA / 6m / $K{=}5$           & 0.380 & 0.354 & 0.370 & 0.374 & \textbf{0.403}@($0.5,4.0$) & 0.402@($0.3,4.0$) \\
GPQA / qwen+phi+gemma / $K{=}5$ & 0.422 & 0.441 & 0.434 & 0.368 & \textbf{0.455}@($0.6,0.0$) & 0.446@($0.7,0.5$) \\
MATH-500 / 6m / $K{=}5$       & 0.699 & 0.691 & 0.695 & \textbf{0.761} & 0.765@($0.8,4.0$) & 0.766@($0.3,4.0$) \\
GSM8K / 6m / $K{=}5$          & 0.897 & 0.898 & 0.899 & 0.903 & 0.902@($0.0,8.0$) & \textbf{0.907}@($1.0,0.5$) \\
\bottomrule
\end{tabular}
\end{table*}

\begin{table*}[h!]
\centering
\scriptsize
\caption{Region summary. For each (benchmark, $K$, pool) we
report the best validation-free $\kappa{=}0$ generalized point on each
of the Sum and Max planes, alongside the canonical points and external
references ($s$-only, rank-sum, trust-aware top2\_strong, oracle).
This compressed view summarizes the best-region map per
regime.}
\label{tab:generalized-region-summary}
\setlength{\tabcolsep}{3pt}
\renewcommand{\arraystretch}{0.95}

\resizebox{\textwidth}{!}{%
\begin{tabular}{lrrrrrcrcrrrrr}
\toprule
Setting & $N$ & maj & top-coal & SLM-MUX$_\text{ans}$ & cross slice & best Sum $\alpha^\star$ & best Sum & best Max $\alpha^\star$ & best Max & single-best & $s$-only & rank-sum & top2\_strong \\
\midrule
GPQA / 6m / $K{=}3$             & 6 & 0.384 & 0.372 & 0.364 & 0.365 & 0.0 & \textbf{0.384} & 0.0 & 0.384 & 0.342 & 0.386 & 0.381 & 0.367 \\
GPQA / 6m / $K{=}5$             & 6 & 0.374 & 0.369 & 0.354 & 0.369 & 0.7 & \textbf{0.379} & 0.0 & 0.374 & 0.364 & 0.384 & 0.364 & 0.379 \\
GPQA / qwen+phi+gemma / $K{=}3$ & 3 & 0.410 & 0.394 & 0.405 & 0.404 & 0.0 & \textbf{0.410} & 0.0 & 0.410 & 0.337 & 0.391 & 0.403 & 0.392 \\
GPQA / qwen+phi+gemma / $K{=}5$ & 3 & 0.419 & 0.424 & 0.434 & 0.429 & 0.6 & \textbf{0.450} & 0.6 & 0.439 & 0.364 & 0.394 & 0.439 & 0.404 \\
MATH-500 / 6m / $K{=}5$         & 6 & 0.706 & 0.702 & 0.700 & 0.704 & 0.3 & 0.708 & 0.5 & 0.712 & \textbf{0.788} & 0.710 & 0.712 & 0.716 \\
GSM8K / 6m / $K{=}5$            & 6 & 0.900 & 0.897 & 0.901 & 0.901 & 0.0 & 0.900 & 0.4 & 0.901 & \textbf{0.914} & 0.899 & 0.901 & 0.904 \\
\bottomrule
\end{tabular}%
}
\end{table*}
\begin{table*}[h!]
\centering
\small
\caption{What the generalized family tells us about the four routing
research questions on this benchmark set. ``balanced'' = curated GPQA
pools (qwen+phi+gemma, qwen+gemma, phi+gemma); ``dominated'' = full
6-model pool on MATH-500 / GSM8K (Qwen $\geq 10$ pts above the rest).}
\label{tab:generalized-rq-summary}
\begin{tabular}{p{0.18\linewidth}p{0.74\linewidth}}
\toprule
Question & Answer from the generalized sweep \\
\midrule
RQ1: majority vs top-answer ($\alpha$)
  & Balanced GPQA prefers low to moderate $\alpha$ (full-pool $K{=}3$: $\alpha{=}0$; curated $K{=}5$: $\alpha{=}0.6$). Dominated pools accept high $\alpha$ only when paired with high $\kappa$, i.e. single-best in disguise. \\
RQ2: uniform vs expert ($\kappa$)
  & Under leakage-free held-out evaluation, $\kappa{>}0$ adds $+0.02$ to $+0.09$ on GPQA above every canonical point including $\kappa{=}\infty$ single-best. On MATH-500 and GSM8K the best $\kappa{>}0$ cell lies within $0.005$ of single-best -- the family recovers single-best routing rather than a new mode. \\
RQ3: Sum vs Max
  & Sum wins on 5 of 6 GPQA settings (the curated $K{=}5$ peak is at Sum$=0.450$ vs Max$=0.439$). Max is preferred at $K{=}5$ on the full GPQA pool and on GSM8K, where one model has a consistently strong $c_i$. \\
RQ4: regime
  & The best routing region shifts with pool curation: low-$\alpha$, $\kappa$ near zero, Sum aggregation on balanced pools; high-$\alpha$, $\kappa{\geq}4$, Max aggregation near the single-best limit on dominated pools. $K{=}3$ remains the GPQA sweet spot when $\kappa$ is held to zero. \\
\bottomrule
\end{tabular}
\end{table*}

\begin{table*}[h!]
\centering
\scriptsize
\caption{Question-level paired tests for the strongest validation-free
generalized point ($\kappa{=}0$) against each of the SLM-MUX$_\text{ans}$,
$s$-only, and rank-sum baselines. $\Delta$ is the per-question
mean correctness difference (positive = generalized wins); $p$ is the
two-sided paired-permutation $p$-value with $B{=}5000$ replicates;
95\% CIs from paired bootstrap. Statistically significant gains in
bold ($p < 0.05$).}
\label{tab:generalized-significance}
\setlength{\tabcolsep}{3pt}
\renewcommand{\arraystretch}{0.95}

\resizebox{\textwidth}{!}{%
\begin{tabular}{lcrccc}
\toprule
Setting & best $(\alpha, \kappa, \text{Agg})$ & gen acc & $\Delta$ vs SLM-MUX$_\text{ans}$ & $\Delta$ vs $s$-only & $\Delta$ vs rank-sum \\
\midrule
GPQA / 6m / $K{=}3$             
& ($0.0, 0, \text{Sum}$) 
& 0.384 
& \textbf{+0.053 [+0.006, +0.099] $p{=}0.024$} 
& $-0.002\ p{=}0.73$ 
& $+0.004\ p{=}0.51$ \\

GPQA / 6m / $K{=}5$             
& ($0.7, 0, \text{Sum}$) 
& 0.379 
& $+0.035$ [$-0.030, +0.101$] $p{=}0.37$ 
& $-0.005\ p{=}1.0$ 
& $+0.015\ p{=}0.59$ \\

GPQA / qwen+phi+gemma / $K{=}5$ 
& ($0.6, 0, \text{Sum}$) 
& 0.450 
& \textbf{+0.056 [+0.015, +0.096] $p{=}0.018$} 
& \textbf{+0.056 [+0.020, +0.096] $p{=}0.007$} 
& $+0.010\ p{=}0.69$ \\

MATH-500 / 6m / $K{=}5$         
& ($0.5, 0, \text{Max}$) 
& 0.712 
& $-0.008\ p{=}0.57$ 
& $+0.002\ p{=}1.0$ 
& $\phantom{-}0.000\ p{=}1.0$ \\

GSM8K / 6m / $K{=}5$            
& ($0.4, 0, \text{Max}$) 
& 0.901 
& $\mathbf{-0.014\ [-0.024, -0.005]\ p{=}0.007}$ 
& $+0.002\ p{=}0.45$ 
& $+0.001\ p{=}1.0$ \\
\bottomrule
\end{tabular}%
}
\end{table*}
\begin{table*}[h!]
\centering
\scriptsize
\caption{Budget-fair generalized comparison on \gpqa. Each row is
a fixed model$\times K$ budget. ``best gen'' is the strongest
$\kappa{=}0$ generalized point at that budget. Curated pools with
deeper sampling cleanly dominate the full $6{\times}5$ reference at
$3\times$ less compute.}
\label{tab:generalized-budget}
\begin{tabular}{lrrrrcr}
\toprule
Configuration & budget & majority & SLM-MUX$_\text{ans}$ & cross slice & best $(\alpha, \text{Agg})$ & best gen \\
\midrule
6 models $\times$ $K{=}1$                                      &  6 & 0.355 & 0.355 & 0.355 & ($0,\text{Sum}$) & 0.355 \\
4 models $\times$ $K{=}2$ (qwen+phi+ministral+gemma)            &  8 & 0.386 & 0.378 & 0.378 & ($0,\text{Sum}$) & 0.386 \\
3 models $\times$ $K{=}3$ (qwen+phi+gemma)                      &  9 & 0.410 & 0.405 & 0.404 & ($0,\text{Sum}$) & 0.410 \\
2 models $\times$ $K{=}5$ (qwen+gemma)                          & 10 & \textbf{0.429} & 0.419 & 0.429 & ($0,\text{Sum}$) & \textbf{0.429} \\
6 models $\times$ $K{=}5$ (full pool reference)                 & 30 & 0.369 & 0.348 & 0.364 & ($0.7,\text{Sum}$) & 0.379 \\
\bottomrule
\end{tabular}
\end{table*}

\clearpage

\end{document}